\documentclass{article}
\pdfoutput=1

\usepackage{arxiv}

\usepackage{microtype}
\usepackage{graphicx}
\usepackage{caption}
\usepackage{subcaption}
\usepackage[utf8]{inputenc} 
\usepackage[T1]{fontenc}    
\usepackage{hyperref}       
\usepackage{url}            
\usepackage{booktabs}       
\usepackage{amsfonts}       
\usepackage{amsmath}

\usepackage{MnSymbol}%
\usepackage{wasysym}%

\usepackage{nicefrac}       
\usepackage{microtype}      
\usepackage{enumitem}
\usepackage{multirow}
\usepackage[table,xcdraw]{xcolor}
\usepackage{algorithm}
\usepackage{algpseudocode}
\usepackage{sidecap}

\title{Efficient model compression with Random Operation Access Specific Tile (ROAST) hashing}

\author{%
  Aditya Desai\thanks{Department of Computer Science, Rice University, Houston, Tx 77005} \\
  \texttt{apd10@rice.edu} \\
  \And 
  Keren Zhou\footnotemark[1] \\
  \texttt{kz21@rice.edu} \\
  \And 
  Anshumali Shrivastava\footnotemark[1] \thanks{ThirdAI Corp. Houston, Texas} \\
  \texttt{as143@rice.edu} \\
}
\newcommand{\ip}[2]{\langle #1,#2 \rangle}
\newcommand{\roach}{ROAST}
\newcommand{\roast}{ROAST}
\newcommand{\roachfull}{Random Operation Access Specific Tile}
\newcommand{\ssection}[1]{\vspace{-0.25cm} \section{#1} \vspace{-0.3cm}}
\newcommand{\ssubsection}[1]{\vspace{-0.25cm} \subsection{#1} \vspace{-0.25cm}}
\newtheorem{theorem}{Theorem}
\begin{document}
\maketitle

\begin{abstract}
Advancements in deep learning are often associated with increasing model sizes. The model size dramatically affects the deployment cost and latency of deep models. For instance, models like BERT cannot be deployed on edge devices and mobiles due to their sheer size. As a result, most advances in Deep Learning are yet to reach the edge. Model compression has sought much-deserved attention in literature across natural language processing, vision, and recommendation domains. This paper proposes a model-agnostic, cache-friendly model compression approach: Random Operation Access Specific Tile (\roach) hashing. 
{\roast} collapses the parameters by clubbing them through a lightweight mapping. Notably, while clubbing these parameters, {\roast} utilizes cache hierarchies by aligning the memory access pattern with the parameter access pattern. 
{\roach} is up to ${\sim}25\times$ faster to train and ${\sim}50\times$ faster to infer than the popular parameter sharing method HashedNet. Additionally, {\roast} introduces global weight sharing, which is empirically and theoretically superior to local weight sharing in HashedNet, and can be of independent interest in itself. With {\roast}, we present the first compressed BERT, which is 100$\times$-1000$\times$ smaller but does not result in quality degradation. These compression levels on universal architecture like transformers are promising for the future of SOTA model deployment on resource-constrained devices like mobile and edge devices. 
\end{abstract}







\ssection{Introduction}
Models across different domains, including Natural Language Processing (NLP), Computer Vision (CV), and Information Retrieval (IR), are exploding in size.
State-of-the-art (SOTA) results in these domains are being obtained at a disproportionate increase in model sizes, questioning the sustainability of deep learning~\cite{thompson2021deep}.
For instance, SOTA architectures for vision include VGG~\cite{vgg} (150M params, 0.6GB) and ViT~\cite{vit} (up to 304M params, 1.2GB). Additionally, SOTA NLP architectures include BERT~\cite{devlin2018bert} (340M params, 1.36GB), GPT-2~\cite{gpt} (1.5B params, 6GB), MegatronLM~\cite{megatron} (8.3B params, 34GB), T5~\cite{t5} (11B params, 44GB), T-NLG~\cite{nlg} (17B params, 68GB), GShard~\cite{gshard} (600B params, 2.4TB) and so on.
Similarly, industrial-scale recommendation models such as DLRM~\cite{DLRM19} can have up to 100 billion parameters.
Models as large as these can cause many issues in various aspects.

Many real-world applications require deploying models on the edge and mobile devices. However, SOTA models cannot be deployed on resource-constrained devices due to their size.
These models are often deployed on high-end servers, and applications on devices need to communicate data to these servers to retrieve the results. This mechanism makes the service slow due to communication time and creates a data privacy concern. Significant model compression can potentially resolve this issue. For instance, BERT or GPT2, widely used SOTA NLP models, can be deployed on edge if compressed up to $100\times$.
Additionally, many NLP and recommendation models cannot be trained on a single GPU, and they require distributed model-parallel training. Model parallel training is time-consuming owing to communication costs. Also, efficient training on such distributed systems often requires engineering expertise. However, $100\times$ compressed models of T-NLG or $1000\times$ compressed models of DLRM or GShard can be easily trained on a single GPU. Additionally, even for any distributed training and particularly federated learning~\cite{konevcny2015federated} communication is directly proportional to model size. Thus, model sizes become the primary bottleneck in such training. Again, $1000\times$ compressed models can obtain a commensurate reduction in communication costs.

Furthermore, compressing large models to small sizes come with immediate latency benefits. For example, \cite{diamos2016persistent} showed that if a single RNN layer can fit in registers, then it leads to $146\times$ faster inference. Also, \cite{robez} showed that by compressing the DLRM model $1000\times$ and using a single GPU instead of 8 GPUs, we could get $3\times$ faster inference at a lower cost. Indeed, accessing RAM is orders of magnitude slower and costlier than performing computation.

Thus, the ML community has heavily invested in model compression. A variety of model compression paradigms now exist in literature like pruning~\cite{Han2016DeepCC}, quantisation~\cite{Han2016DeepCC}, knowledge distillation~\cite{modelcompression}, parameter-sharing~\cite{hashtrick,robez}, and low rank decomposition~\cite{hrinchuk2020tensorized, ttrec}.
We will discuss each of these paradigms in Section \ref{sec:related}. This paper follows the parameter-sharing world of model compression.
Also, we will focus on the NLP compression for the sake of discussion in this paper. However, it should be noted that our proposed model compression technique is model agnostic and can be applied to any model in any domain.

Parameter-sharing methods use a small repository of weights shared among the various parameters of the model. There are various methods in parameter-sharing, which vary depending on the sharing scheme. For example, some compressed transformer architectures such as ALBERT~\cite{lan2019albert} and UT~\cite{dehghani2018universal} apply the same layer multiple times. HashedNet~\cite{hashtrick}, ROBE-Z~\cite{robez} and SlimEmbeddings~\cite{li2018slim} share parameters using random mappings. Most parameter-sharing techniques in literature are devised for a specific model component. This paper introduces {\roachfull} (\roach) hashing, a parameter-sharing approach, which can be applied to all computational modules of a model such as MLP layers, attention blocks, convolution layers, and embedding tables. {\roast} leverages a recent result in hashing, which shows that chunk-based hashing is theoretically superior to usual feature hashing~\cite{robez}. {\roach} proposes a tile-based hashing scheme that is tuned to the memory access pattern of the algorithmic implementation of the operation being performed. Thus, {\roast} gives us an efficient knob on the memory footprint of the model without affecting its functional form.
Additionally, {\roach} also proposes global weight-sharing where parameters are shared across the different computational modules. As we shall see, global weight-sharing is both empirically and theoretically superior to local weight sharing and might be of independent interest.  

We evaluate {\roach} compression on the BERT, one of the popular NLP models with transformer architecture. Transformers are emerging as a universal architecture showing good results in all the three domains, NLP~\cite{devlin2018bert,vaswani2017attention}, CV~\cite{vit} and IR~\cite{moreira2021transformers}. With {\roach}, we could compress the BERT model as much as 100$\times$-1000$\times$ without any loss of quality in training-from-scratch on the text classification task. This is orders of magnitude larger than previously known BERT compression results in NLP. To the best of our knowledge, most BERT model compression beyond 2$\times$-3$\times$ have shown degradation in model quality \cite{gupta2022compression}. What makes this result exciting is that a $100\times$ {\roach}ed BERT model is small enough to fit the cache of CPUs and is easily deployable on mobile and edge devices. Additionally, {\roast} is much faster than competing approaches like HashedNet. With {\roast}, inference can be up to ${\sim}50\times$ faster whereas training can be ${\sim}25\times$ faster than HashedNet.

\textbf{Limitations of {\roach}:} One of the goals of model compression, apart from reducing memory usage, is to reduce computational workload for deployment. {\roach}, currently, is not optimized enough to decrease computation; it only decreases the memory footprint of a model. Reducing computation with a small memory is left for future work. However, it should be noted that reducing the memory footprint itself can reduce computation latency and power consumption significantly.
As shown in \cite{han2016eie}, accessing memory from RAM is 6400$\times$ costlier than 32bit INT ADD and 128$\times$ costlier than on-chip SRAM access in terms of energy consumption. Additionally, RAM access generally is ${\sim}100 \times$ slower than a floating-point operation. This shows that reducing memory footprint so that model can fit in faster memory can, at times, be much more impactful than reducing the computational burden.

\ssection{Related Work}\label{sec:related}
There is a rich history of model compression in various domains such as NLP, CV, and IR. Model compression can be generally classified into two categories: (1) Compressing a learned model and (2) Learning a compressed model. {\roach} lies in the second category. This section briefly reviews general paradigms in model compression. We keep the discussion in the context of NLP models and occasionally mention results on BERT compression. For a comprehensive survey on NLP model compression, we refer readers to the survey~\cite{gupta2022compression}.

\textbf{Compressing learned models:} 
\textbf{1) Pruning:} Pruning is a technique to remove parts of a large model, including weights, nodes, blocks, and layers, to make the model lighter. Pruning can be performed as a one-time operation or gradually interspersed with training. ~\cite{guo2019reweighted,fan2019reducing} showed 2$\times$-3$\times$ compression on the BERT model on certain textual entailment, question answering, and sentiment analysis datasets with similar or better quality. 
\textbf{2) Quantization:} Quantization can involve reducing the precision of the parameters of a model. Mixed precision models are sometimes used where different precision is used with different weights. Another way to quantize is KMeans quantization, where weights of the models are clustered using KMeans, and each cluster's centroid replaces a quantized weight. Product quantization~\cite{jegou2010product} is a particular type of KMeans quantization. 
\cite{shen2020q} showed ${\sim}16\times$ compression with mixed precision quantization on BERT. However, the compression yields a significantly worse model quality on textual entailment, question answering, and named entity recognition task.
\textbf{3) Knowledge distillation: } Knowledge distillation \cite{modelcompression} is widely applied in NLP model compression with a focus on distilled architectures. Knowledge distillation involves first training a teacher model; then, a student model is trained using logits of the teacher model. Many variations exist on this basic idea of knowledge distillation. An example in literature for BERT distillation is DistilBERT \cite{sanh2019distilbert} with 2.25$\times$ compression and similar or better quality.

While these techniques have been widely successful, one of the drawbacks of this line of compression is the need to first have a trained large model, which is then compressed.


\textbf{Learning compressed models}
\textbf{1) Low-rank decomposition:} Under the low-rank assumption, a matrix can be represented as a product of two low-rank matrices. This technique is often used in reducing the model memory by decomposing large matrices. Tensor-train (TT) decomposition is a generalization of low-rank decomposition applied to tensors. TT-embeddings~\cite{hrinchuk2020tensorized} is an example of TT decomposition applied to embeddings in NLP.
\textbf{2) Parameter sharing:} Parameter sharing approaches such as HashedNet~\cite{hashtrick} and SlimEmbeddings~\cite{li2018slim} are primarily applied to embedding tables.
These approaches randomly share individual parameters or sub-vectors (in the case of slim embeddings). Character-aware language models~\cite{kim2016character} build word embeddings from character embeddings, thus reducing the embedding table size. Another kind of parameter sharing is used in transformer models such as ALBERT~\cite{lan2019albert} and Universal Transformer~\cite{dehghani2018universal} in which weights are repeatedly used in subsequent layers. This leads to ${\sim}18\times$ compression but with degradation of quality. {\roast} follows the model-agnostic parameter-sharing style of model compression research.
\ssection{Background}

\textbf{HashedNet: Compressing MLP matrices}
Previous work~\cite{hashtrick} introduced a weight sharing method to compress weight matrices of MLP models.
They map each matrix parameter to a shared parameter array using a random hash function xxhash~\cite{collet2016xxhash}.
In the forward pass, this mapping is used to recover a weight matrix and perform matrix multiplication for each MLP layer.
In the backward pass, the gradients of each weight matrix are mapped to the shared compressed array and aggregated using the sum operation.
It should also be noted that each MLP layer uses an independent array of parameters.
One of the main concerns with HashedNet is that memory accesses on the compressed array are non-coalesced.
Thus, fetching a compressed matrix via HashedNet requires significantly more memory read transactions than fetching an uncompressed matrix for which memory accesses can coalesce.
Our evaluation shows that uncoalesced memory accesses lead to high latency, especially for large matrices.

\textbf{Random Block Offset Embedding Array (ROBE) for embedding compression}
In ROBE~\cite{robez}, the embedding table is generated using an array of parameters.
The embedding of a token is obtained by drawing chunks of the embedding from the ROBE array.
The locations of the chunks are decided randomly via light-weight universal hash functions.
Authors of ROBE showed that ROBE hashing is theoretically superior to feature hashing used in HashedNet.
Also, the use of chunks causes memory accesses to coalesce, making embedding lookup efficient.

\ssection{{\roachfull} (\roach) hashing}
\begin{figure}
    \centering
    \includegraphics[scale=0.25]{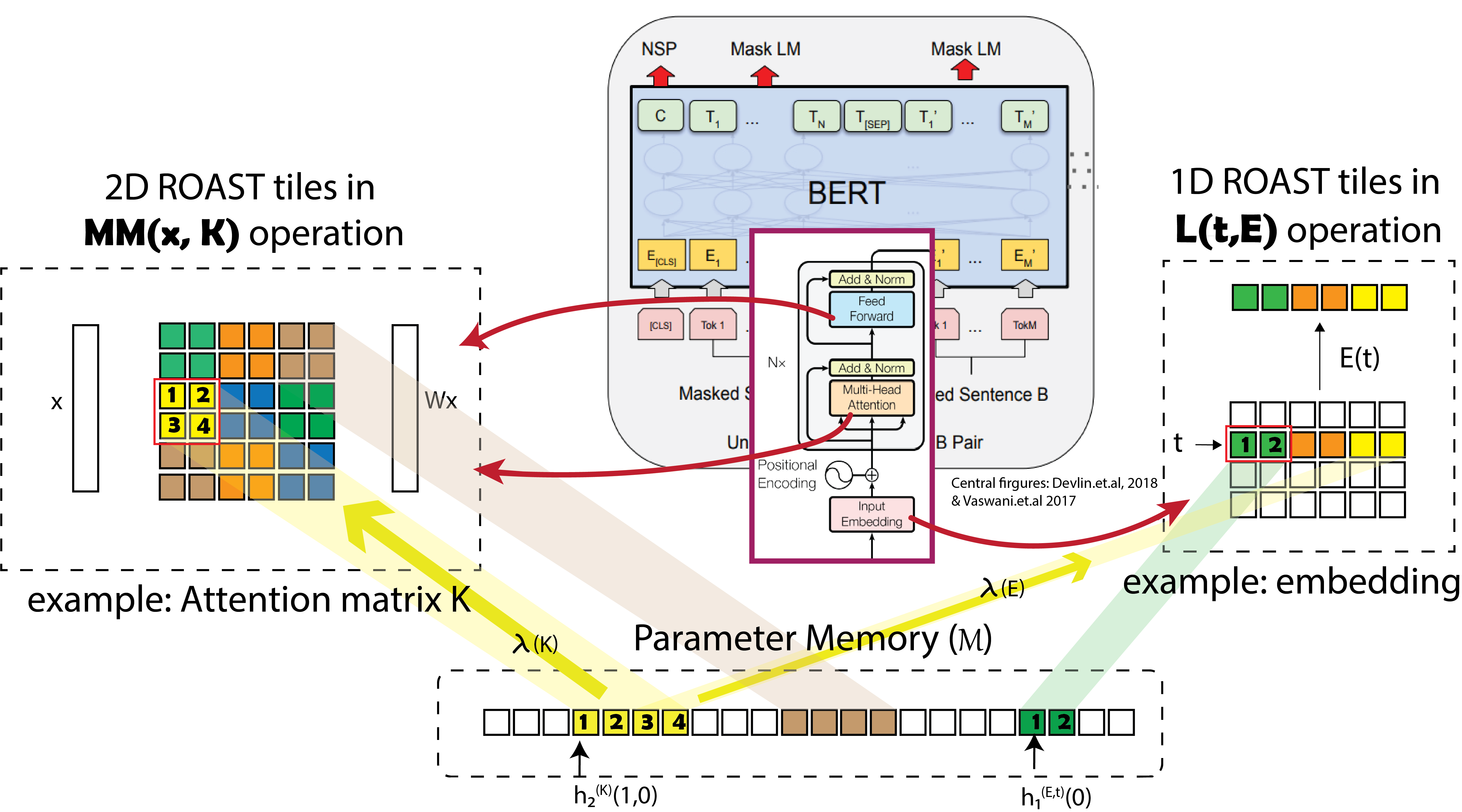}
    \caption{Generic model compression with operation-specific blocking for BERT as an example : (left) Shows how 2D tiles are mapped to $\mathcal{M}$ in case of $\mathbf{MM}$ operation. (right) Shows how 1D tiles are mapped to $\mathcal{M}$ in case of $\mathbf{L}$ operation. $\lambda$ is the module-specific GMS scaling factor}
    \label{fig:mc}
\end{figure}

Let $\mathcal{M}$ be the compressed memory from which parameters will be used, $f$ be the model or the function that we want to run using $\mathcal{M}$, and $W$ be the recovered weights used in $f$. $f$ can be considered as a composition of operations $\{\mathcal{O}_i(X_i, W_i) \}$. By operation, we mean the smaller functions that, when composed together, give us the model $f$. Here $X_i$ is the input to the operation, and $W_i$ is the weights (i.e., learnable parameters) that $\mathcal{O}_i$ uses. Generally, $W_i$s are distinct and do not share parameters.

Random Operation Access Specific Tile ({\roast}) hashing is a way to perform efficient model-agnostic parameter sharing-based compression. The following distinct aspects of {\roast} set it apart from previous parameter sharing-based methods. (1) {\roast} is a generic technique applicable to all computational modules. (2) {\roast} proposes to tune its mapping from $W_i$ to $\mathcal{M}$ in a way that coalesces memory accesses according to how memory is accessed during the operation. This makes {\roast} efficient and up to $45\times$ faster than competing approaches like HashedNet. (3) {\roast} proposes Global Memory Sharing (GMS) as opposed to Local Memory Sharing (LMS) used in HashedNet. We show GMS to be theoretically and empirically superior to LMS in Section~\ref{sec:theory} and~\ref{sec:robez_exp}.

\ssubsection{{\roast} operations in deep learning} \label{sec:operations}
Any model $f$ can be considered as a composition of smaller functions $\{\mathcal{O}_i(X_i, W_i) \}$. There are multiple ways to perform this decomposition depending upon what we consider a valid (or small enough) operation. In {\roast}, we consider three types of operations: (1) $\mathbf{L}(l, W)$, lookup that accesses $\mathcal{M}$ and recovers $l^{th}$ element of $W$, say $w$. By element, we mean some particular part of $W$ that is identifiable by an integer. An example with embedding tables is given in figure \ref{fig:mc}. (2) $\mathbf{MM}(X, W)$, matrix multiplication that multiplies $X$ with $W$ and returns the result, and (3) $\mathbf{N}(X)$, various operations that only act on the input but do not interact with $\mathcal{M}$. 
In {\roast}, in order to limit the memory usage, we make sure that $\mathbf{L}$ is used only on a small $w$ and $\mathbf{MM}$ is performed without recovering the entire matrix.
We find that most deep learning models, if not all, can be written as a composition of operations $\mathbf{N}$, $\mathbf{MM}$ and $\mathbf{L}$, where $\mathbf{L}$ is only applied on small parameters.
Let us discuss how {\roast} implements $\mathbf{L}$ and $\mathbf{MM}$ operations in the following paragraphs.

\textbf{Lookup ($\mathbf{L}(l, W)$)} We recover a parameter weight $w$ of any shape in a row-major format. Thus, we can consider $w=W(l)$ to be a 1D vector without loss of generality. {\roast} recovers $w$ from $\mathcal{M}$ in a blocked fashion. Consider $w$ to be composed of chunks of size $Z$. Each chunk $c$ is located in $\mathcal{M}$ using a universal hash function $h_1$ and is recovered from the location $h_1(c)$ in $\mathcal{M}$. Let $C(i)$ give the chunk number of index $i$ and $O(i)$ give the offset of $i$ in this chunk. 
\begin{align}
    w[i] = \lambda \mathcal{M}[h_1(C(i)) + O(i)] \quad \quad h_1 : \mathbb{N} \rightarrow \{0,..., |\mathcal{M}| - Z\}
\end{align}
The recovered $W$ has $\lambda$ as a scaling factor discussed in section \ref{sec:gms}.
The hash function hashes to a range $\{0,..., |\mathcal{M}| - Z\}$ to avoid overflows while reading the memory.
For example, Figure \ref{fig:mc} (right) illustrates the embedding lookup using $\mathbf{L}$ with chunk size of 2.
{\roast} uses $\mathbf{L}$ to implement computational modules such as embeddings, bias vectors, and so on.
We generalize the embedding lookup kernel from ROBE~\cite{robez} to implement our $\mathbf{L}$ kernel. 

\textbf{Matrix multiplication ($\mathbf{MM}(X_i, W_i)$)} 2D matrix multiplication is one of the most widely used operations in deep learning.
We implement our \roast-MM kernel with parameter sharing performed in a way that the algorithm for matrix multiplication accesses coalesced pieces of $\mathcal{M}$.
An efficient implementation of matrix multiplication on GPU follows a block multiplication algorithm to use the on-chip shared memory efficiently. While computing $C=A \times B$, A, B and C are divided in tiles of size $Z_0\times Z_1$, $Z_1\times Z_2$ and $Z_0\times Z_2$ respectively. 
Thus, we divide our 2D weight matrix into tiles of size $Z_1 \times Z_2$.
The tile, $(x,y)$, where $x$ and $y$ are the coordinates of the tile, is located in $\mathcal{M}$ in a row-major format via a universal hash function $h_2(x,y)$. Let $C_1(i,j)$ and $C_2(i,j)$ give the $x$-coordinate and $y$-coordinate of the tile to which $i$, $j$ belongs. Similarly, let $O_1(i,j)$ and $O_2(i,j)$ give the $x$-offset and $y$-offset of a location $(i,j)$ on the tile. Then, we use the following mapping for \roast-MM,
\begin{align*}
    W[i,j] = \lambda &\mathcal{M}[h_2(C_1(i,j), C_2(i,j)) + Z_2 O_1(i,j) + O_2(i,j)] \\
    &h_2 : \mathbb{N}^2 \rightarrow \{0,...,|\mathcal{M}|-Z_1Z_2\}
\end{align*}
Again, $\lambda$ is the scaling factor discussed in section \ref{sec:gms}. The hash function hashes to a range $\{0, ..., |\mathcal{M}| - Z_1Z_2\}$ to avoid overflows while reading the chunk.
Figure \ref{fig:mc} (left) illustrates \roast-MM with a chunk size of $2\times2$.
The above mapping is used whenever a 2D tile is accessed in the matrix multiplication algorithm.
The pseudo code for \roast-MM is shown in algorithm \ref{algo:roach-mm}. Unfortunately, existing optimized libraries for matrix multiplication such as CUTLASS\cite{cutlass} or cuBLAS\cite{cublas} do not support custom tile loading. Hence, we implement our own \roast-MM kernel in Triton \cite{triton}.
More details on our implementation and its performance are presented in Section~\ref{sec:triton}.
{\roast} uses \roast-MM kernel to implement computational modules such as MLP layers, attention blocks, etc.
Each module invoking {\roast} kernels uses independent hash functions.
\begin{algorithm}
\caption{\roach-MM($I \times H \times O$)}\label{algo:roach-mm}
\begin{algorithmic}
\Require $X\in R^{I \times H}$, $\mathcal{M}$, $\lambda$, $h:\mathbb{N}^2 \rightarrow \{0,...,|\mathcal{M}|-Z_1Z_2\}$
\Ensure $output=\mathbf{MM}(X, \mathcal{M}[h(:,:)])$ 
\State $value \leftarrow \mathbf{TILE}(Z_0, Z_2)$\Comment{Allocate a 2D tile of size $Z_0 \times Z_2$ to accumulate results}
\For{$i \in \{0, 1,  ..., \lceil I / Z_0 \rceil-1\}$ }
     \For{$j \in \{0, 1,  ..., \lceil O / Z_2 \rceil-1\}$}
            \State $value[:,:] \leftarrow 0$
            \For{$k \in \{0, 1,  ..., \lceil H / Z_1 \rceil-1\}$}
                \State $value \leftarrow value +  \mathbf{MM}(X[i:i+Z_0,k:k+Z_1], \mathcal{M}(h(k:k+Z_1, j:j+Z_2)))$ 
                
                \Comment{Access to the weight tile passes through the hash function}
            \EndFor
            \State $output[i:i+Z_0,j:j+Z_2] \leftarrow \lambda * value$
        \EndFor
\EndFor 

\end{algorithmic}
\end{algorithm}

Apart from scaling each recovered parameter with module-specifc $\lambda$, we can also multiply it with another independent hash function $g:\mathbb{N}^k \rightarrow \{\pm 1\}$ ($k$=1 or $k$=2).

\ssubsection{Global memory sharing (GMS)}\label{sec:gms}

HashedNet uses local memory sharing (LMS), which states that each layer will have independent compressed memory.
In contrast, {\roach} proposes global memory sharing (GMS), wherein we share memory across modules.
However, modules cannot directly use the parameters stored in $\mathcal{M}$ as each module's weights requires initialization and optimization at different scales.
For instance, in the Xavier's initialization \cite{glorot2010understanding}, weights are initialized with distribution $\mathbf{Uniform}(-1/\sqrt{n}, 1/\sqrt{n})$ where $n$ is size of the input to the module.
In GMS, we must ensure that each module gets weights at the required scale.
To achieve this, we first initialize the entire {\roach} parameter array with values from the distribution $\mathbf{Uniform}(-1/C,1/C)$ for some constant $C$.
Then, for each module, we scale the weights retrieved from the {\roach} array by a factor of $\lambda = C/\sqrt{n}$.

One can understand the benefit of GMS over LMS in terms of the number of distinct functions in $f$ that can be expressed using a fixed $\mathcal{M}$.
Consider a family of functions with $n$ parameters.
GMS can potentially express $|\mathcal{M}|^n$ functions across different random mappings. In LMS, let separate parameters be of sizes ${n_1, n_2, .. n_k}$ and each of them is mapped into memories ${\mathcal{M}_1, \mathcal{M}_2,...,\mathcal{M}_k}$. Thus, $n=\sum_i n_i$ and $|\mathcal{M}| = \sum_i |\mathcal{M}_i|$. Then LMS can only express $|\mathcal{M}_1|^{n_1}|\mathcal{M}_2|^{n_2}....|\mathcal{M}_k|^{n_k}$  different functions. Thus expressivity of LMS is strictly less than that of GMS and can be orders of magnitude less depending on exact values of $n_i$ and $|\mathcal{M}_i|$. We also show that GMS is superior to LMS in terms of dimensionality reduction (feature hashing) in Section~\ref{sec:theory}.

\ssubsection{Forward and backward passes}
Recall that in {\roast}, operations are of three types $\mathbf{L}, \mathbf{MM}$ and $\mathbf{N}$.
The forward pass proceeds by applying each operation in sequence.
If an operation is of type $\mathbf{N}$, we directly apply its function on the input.
For $\mathbf{L}$ and $\mathbf{MM}$ operations, outputs are computed according to the procedure described in Section~\ref{sec:operations}.

The gradient of the loss w.r.t a weight in $\mathcal{M}$ is the $\lambda$-scaled aggregation of gradients of loss w.r.t all the parameters that map to this weight. For simplicity of notation, consider $\theta$ as the complete parameter, $\lambda(j)$ as the scaling factor we use for the module that $\theta_j$ belongs to, and $h$ be a hash function.
\begin{equation}
    \nabla_{w_i} f(w) = \sum_{j , h(j) = i} \lambda(j) * \nabla_{\theta_j} f(\theta)
\end{equation}
This is because
\begin{align}
    \frac{\partial f(x, g(x))}{\partial x} = \frac{\partial f(z, g(y))}{\partial z}|_{y=x, z=x} + \frac{\partial f(z, g(y))}{\partial y} |_{y=x, z=x}\label{eq:1}
\end{align}
Equation \ref{eq:1} shows that for gradient computation, we can rename the variables from different modules, which are mapped to a single weight, and compute gradients w.r.t renamed variables. The gradient w.r.t to the weight is just the sum of the individual gradients. See Appendix \ref{sec:gradexp} for details.

\ssection{Feature hashing quality: global memory sharing advantage over local memory sharing} \label{sec:theory}
We can consider model compression as dimensionality reduction of a parameter vector (a one dimensional vector of all parameters in a model) of size $n$ into a vector of size $|\mathcal{M}|=m$. Quality of inner-product preservation is used as a metric to measure the quality of dimensionality reduction.
In terms of dimensionality reduction, {\roach} uses ROBE hashing, which shows that chunk based hashing is theoretically better than hashing individual elements.
In this section, we compare {\roach}'s GMS proposal against HashedNet's LMS using a chunck size of one.
Consider two parameter vectors $x, y \in R^n$, we are interested in how the inner product of parameter vectors are preserved under hashing.
Let $x = [x_1, x_2, ..., x_k]$ and $y = [y_1, y_2, ..., y_k]$ be composed of $k$ vectors of sizes $n_1, n_2,...n_k$ where [] denotes concatentation.
In LMS, let each piece map to memory of size $f_i m$ where $\sum_{i} f_i = 1$.
The estimated inner product with GMS is
\begin{equation}
    \widehat{\ip{x}{y}}_{G,m} = \sum_{j=1}^m \left(\sum_{i=1}^n \mathbb{I}(h(i){=}j) g(i) x[i] \sum_{i=1}^n \mathbb{I}(h(i){=}j) g(i) y[i]\right)
\end{equation}
The estimated inner product with LMS can be written as
\begin{equation}
    \widehat{\ip{x}{y}}_{L,m,\Vec{f}} = \sum_{l=1}^k \sum_{j=1}^{f_l m} \left(\sum_{i=1}^{n_l} \mathbb{I}(h(i){=}j) g(i) x_l[i] \sum_{j=1}^{n_l} \mathbb{I}(h(i){=}j) g(i) y_l[i]\right) = \sum_{l=1}^k \widehat{\ip{x_l}{y_l}}_{G,(f_l m)}
\end{equation}
\begin{theorem}
Let $x,y\in R^n$ and be composed of $k$ vectors $x=[x_1, x_2, ..., x_k]$ and $y=[y_1, y_2, ..., y_k]$. Then the inner product estimation of global and local weight sharing are unbiased. 
\begin{equation}
    \mathbb{E}(\widehat{\ip{x}{y}}_{G,m}) = \ip{x}{y} \;\quad \; \mathbb{E}(\widehat{\ip{x}{y}}_{L,m,\Vec{f}}) = \ip{x}{y}
\end{equation}
The variance for inner product estimation can be written as,
\begin{equation}
\mathbb{V}_G (\widehat{\ip{x}{y}}) = \sum_i f_i V_i + \frac{1}{m} \left(\sum_{i,j,i\neq j} (||x_i||^2 ||y_j||^2) + \ip{x_i}{y_i} \ip{x_j}{y_j} \right)
\end{equation}
\begin{equation}
\mathbb{V}_L (\hat{\ip{x}{y}}) = \sum_i V_i
\end{equation}
where 
\begin{equation}
    V_l = \frac{1}{f_l} \frac{1}{m} \left(\sum_{i\neq j} a_i^2 b_j^2  + \sum_{i \neq j} a_i b_i a_j b_j\right) \textrm{, where } x_l = (a_1, a_2 ..., a_{n_l}) \textrm{ and } y_l = (b_1, b_2 ..., b_{n_l})
\end{equation}
where $\mathbb{V}_L$ is local memory sharing variance and $\mathbb{V}_G$ is global memory sharing variance. 
\end{theorem}
\textbf{Intuition:} The two terms in $\mathbb{V}_G$ can be understood as follows: The first term is the local variance with individual terms reduced by a factor of $f_i$. This is because each piece of the vector is being distributed in a memory that is $1/f_i\times$ larger. However, in GMS, there is a possibility of more collisions across pieces. This leads to the second term in $\mathbb{V}_G$. Note that, for a given $x,y$ and a finite value for $m$, $\mathbb{V}_G$ is always bounded. 
At the same time, $\mathbb{V}_L$ is unbounded due to $0<f_i<1$ in the denominator.
So if the number of pieces increases or particular $f_i$ grows smaller, $\mathbb{V}_L$ increases.
While we cannot prove that $\mathbb{V}_G$ is strictly less than $\mathbb{V}_L$, we can investigate the equation under some assumptions on the data.
Practically, each piece of the parameter vector is a computational block like a matrix for multiplication or embedding table lookup.
These blocks are initialized at a scale proportional to the square root of their size.
So the norms of these vectors are similar.
Let us assume the norm of each piece to be $\sqrt{\alpha}$. Also, let us assume that over random data distributions over $x$ and $y$, all the inner products to be $\beta$ in expectation. Then, 
\begin{equation}
    \mathbb{V}_G \approx \frac{k^2}{m} (\alpha^2 + \beta^2) \quad \quad \mathbb{V}_L \approx \frac{1}{m}(\alpha^2 + \beta^2) (\frac{1}{f_1} + \frac{1}{f_2} + ... + \frac{1}{f_k}) \geq  \frac{1}{m}(\alpha^2 + \beta^2) k^2 \frac{1}{(\sum f_i)} = \mathbb{V}_{G}
\end{equation}
Thus, $V_L$ is greater than $V_G$, and it can be much greater depending on the exact values of $f_i$. The proof of the theorem and other details are presented in Appendix~\ref{sec:theory_appx}

\ssection{Implementation and efficiency evaluation of \roach-MM} \label{sec:triton}
\begin{table}[]
\caption{Inference times of different square weight matrices using an input batch of 512. For ROAST, the tile parameters of each matrix multiplication are autotuned. The measurements were taken using TF32 on a NVIDIA A100 GPU (48GB). We used PyTorch's matmul function (MM) for the full uncompressed matrix multiplication. {\color{red}$\blacksquare$}:bad {\color{blue}$\blacksquare$}: good}
\resizebox{\linewidth}{!}{
\begin{tabular}{|cccccccccc|}
\hline
\rowcolor[HTML]{F4F6F8} 
\multicolumn{10}{|c|}{\cellcolor[HTML]{F4F6F8}\textbf{Inference time (ms)}}                                                                                                                                                                                                                                                                                                                                                                                                                                                                                                                                                                                                                                                                                      \\ \hline
\rowcolor[HTML]{F4F6F8} 
\multicolumn{1}{|c|}{\cellcolor[HTML]{F4F6F8}\textit{}}                    & \multicolumn{1}{c|}{\cellcolor[HTML]{F4F6F8}}                                                       & \multicolumn{8}{c|}{\cellcolor[HTML]{F4F6F8}Weight matrix dimensions (Dim $\times$ Dim)}                                                                                                                                                                                                                                                                                                                                                                                                                                                                                                         \\ \hline
\rowcolor[HTML]{F4F6F8} 
\multicolumn{1}{|c|}{\cellcolor[HTML]{F4F6F8}{\color[HTML]{333333} Model}} & \multicolumn{1}{c|}{\cellcolor[HTML]{F4F6F8}{\color[HTML]{333333} $\mathcal{M}$ size $\downarrow$}} & \multicolumn{1}{c|}{\cellcolor[HTML]{F4F6F8}{\color[HTML]{333333} 512}} & \multicolumn{1}{c|}{\cellcolor[HTML]{F4F6F8}{\color[HTML]{333333} 1024}} & \multicolumn{1}{c|}{\cellcolor[HTML]{F4F6F8}{\color[HTML]{333333} 2048}} & \multicolumn{1}{c|}{\cellcolor[HTML]{F4F6F8}{\color[HTML]{333333} 4096}} & \multicolumn{1}{c|}{\cellcolor[HTML]{F4F6F8}{\color[HTML]{333333} 8096}} & \multicolumn{1}{c|}{\cellcolor[HTML]{F4F6F8}{\color[HTML]{333333} 10240}} & \multicolumn{1}{c|}{\cellcolor[HTML]{F4F6F8}{\color[HTML]{333333} 20480}} & {\color[HTML]{333333} Average} \\ \hline
\rowcolor[HTML]{F4F6F8} 
\multicolumn{1}{|c|}{\cellcolor[HTML]{F4F6F8}Full size $\rightarrow$}      & \multicolumn{1}{c|}{\cellcolor[HTML]{F4F6F8}}                                                       & \multicolumn{1}{c|}{\cellcolor[HTML]{F4F6F8}1MB}                        & \multicolumn{1}{c|}{\cellcolor[HTML]{F4F6F8}4MB}                         & \multicolumn{1}{c|}{\cellcolor[HTML]{F4F6F8}16MB}                        & \multicolumn{1}{c|}{\cellcolor[HTML]{F4F6F8}64MB}                        & \multicolumn{1}{c|}{\cellcolor[HTML]{F4F6F8}128MB}                       & \multicolumn{1}{c|}{\cellcolor[HTML]{F4F6F8}420MB}                        & \multicolumn{1}{c|}{\cellcolor[HTML]{F4F6F8}1.6GB}                        &                                \\ \hline
\multicolumn{1}{|c|}{\cellcolor[HTML]{F4F6F8}PyTorch-MM}                   & \multicolumn{1}{c|}{\cellcolor[HTML]{F4F6F8}}                                                       & \multicolumn{1}{c|}{\cellcolor[HTML]{3D85C6}0.10}                       & \multicolumn{1}{c|}{\cellcolor[HTML]{3D85C6}0.11}                        & \multicolumn{1}{c|}{\cellcolor[HTML]{4087C7}0.12}                        & \multicolumn{1}{c|}{\cellcolor[HTML]{5292CC}0.22}                        & \multicolumn{1}{c|}{\cellcolor[HTML]{A5C6E4}0.69}                        & \multicolumn{1}{c|}{\cellcolor[HTML]{FCFDFE}1.18}                         & \multicolumn{1}{c|}{\cellcolor[HTML]{FFFDFD}3.91}                         & \cellcolor[HTML]{CBDEEF}0.91   \\ \hline
\multicolumn{1}{|c|}{\cellcolor[HTML]{F4F6F8}}                             & \multicolumn{1}{c|}{\cellcolor[HTML]{F4F6F8}4MB}                                                    & \multicolumn{1}{c|}{\cellcolor[HTML]{619BD0}0.31}                       & \multicolumn{1}{c|}{\cellcolor[HTML]{669FD2}0.34}                        & \multicolumn{1}{c|}{\cellcolor[HTML]{99BFE1}0.63}                        & \multicolumn{1}{c|}{\cellcolor[HTML]{FFFFFF}2.02}                        & \multicolumn{1}{c|}{\cellcolor[HTML]{FFFBFB}6.20}                        & \multicolumn{1}{c|}{\cellcolor[HTML]{FFF8F8}9.67}                         & \multicolumn{1}{c|}{\cellcolor[HTML]{FFDFDF}35.22}                        & \cellcolor[HTML]{FFF9F9}7.77   \\ \cline{2-10} 
\multicolumn{1}{|c|}{\cellcolor[HTML]{F4F6F8}}                             & \multicolumn{1}{c|}{\cellcolor[HTML]{F4F6F8}32MB}                                                   & \multicolumn{1}{c|}{\cellcolor[HTML]{619BD0}0.31}                       & \multicolumn{1}{c|}{\cellcolor[HTML]{73A7D5}0.41}                        & \multicolumn{1}{c|}{\cellcolor[HTML]{C3D9ED}0.86}                        & \multicolumn{1}{c|}{\cellcolor[HTML]{FFFDFD}3.64}                        & \multicolumn{1}{c|}{\cellcolor[HTML]{FFF4F4}13.66}                       & \multicolumn{1}{c|}{\cellcolor[HTML]{FFECEC}22.11}                        & \multicolumn{1}{c|}{\cellcolor[HTML]{FFAAAA}92.40}                        & \cellcolor[HTML]{FFEFEF}19.06  \\ \cline{2-10} 
\multicolumn{1}{|c|}{\cellcolor[HTML]{F4F6F8}}                             & \multicolumn{1}{c|}{\cellcolor[HTML]{F4F6F8}64MB}                                                   & \multicolumn{1}{c|}{\cellcolor[HTML]{609BD0}0.31}                       & \multicolumn{1}{c|}{\cellcolor[HTML]{7BACD8}0.46}                        & \multicolumn{1}{c|}{\cellcolor[HTML]{EAF2F9}1.09}                        & \multicolumn{1}{c|}{\cellcolor[HTML]{FFFBFB}6.47}                        & \multicolumn{1}{c|}{\cellcolor[HTML]{FFE3E3}31.21}                       & \multicolumn{1}{c|}{\cellcolor[HTML]{FFD9D9}42.45}                        & \multicolumn{1}{c|}{\cellcolor[HTML]{FF5959}178.07}                       & \cellcolor[HTML]{FFDEDE}37.15  \\ \cline{2-10} 
\multicolumn{1}{|c|}{\cellcolor[HTML]{F4F6F8}}                             & \multicolumn{1}{c|}{\cellcolor[HTML]{F4F6F8}128MB}                                                  & \multicolumn{1}{c|}{\cellcolor[HTML]{609BD0}0.31}                       & \multicolumn{1}{c|}{\cellcolor[HTML]{94BBDF}0.60}                        & \multicolumn{1}{c|}{\cellcolor[HTML]{FFFFFF}1.62}                        & \multicolumn{1}{c|}{\cellcolor[HTML]{FFF8F8}9.10}                        & \multicolumn{1}{c|}{\cellcolor[HTML]{FFE0E0}34.62}                       & \multicolumn{1}{c|}{\cellcolor[HTML]{FFCCCC}56.03}                        & \multicolumn{1}{c|}{\cellcolor[HTML]{FF2929}229.31}                       & \cellcolor[HTML]{FFD4D4}47.37  \\ \cline{2-10} 
\multicolumn{1}{|c|}{\cellcolor[HTML]{F4F6F8}}                             & \multicolumn{1}{c|}{\cellcolor[HTML]{F4F6F8}256MB}                                                  & \multicolumn{1}{c|}{\cellcolor[HTML]{629CD1}0.32}                       & \multicolumn{1}{c|}{\cellcolor[HTML]{97BEE0}0.62}                        & \multicolumn{1}{c|}{\cellcolor[HTML]{FFFFFF}1.82}                        & \multicolumn{1}{c|}{\cellcolor[HTML]{FFF7F7}10.25}                       & \multicolumn{1}{c|}{\cellcolor[HTML]{FFDDDD}38.28}                       & \multicolumn{1}{c|}{\cellcolor[HTML]{FFC6C6}62.67}                        & \multicolumn{1}{c|}{\cellcolor[HTML]{FF1010}256.22}                       & \cellcolor[HTML]{FFCFCF}52.88  \\ \cline{2-10} 
\multicolumn{1}{|c|}{\multirow{-6}{*}{\cellcolor[HTML]{F4F6F8}HashedNet}}  & \multicolumn{1}{c|}{\cellcolor[HTML]{F4F6F8}512MB}                                                  & \multicolumn{1}{c|}{\cellcolor[HTML]{649DD1}0.33}                       & \multicolumn{1}{c|}{\cellcolor[HTML]{A3C5E4}0.68}                        & \multicolumn{1}{c|}{\cellcolor[HTML]{FFFFFF}2.05}                        & \multicolumn{1}{c|}{\cellcolor[HTML]{FFF7F7}10.59}                       & \multicolumn{1}{c|}{\cellcolor[HTML]{FFDADA}40.55}                       & \multicolumn{1}{c|}{\cellcolor[HTML]{FFC3C3}65.74}                        & \multicolumn{1}{c|}{\cellcolor[HTML]{FF0000}272.23}                       & \cellcolor[HTML]{FFCCCC}56.03  \\ \hline
\multicolumn{1}{|c|}{\cellcolor[HTML]{F4F6F8}}                             & \multicolumn{1}{c|}{\cellcolor[HTML]{F4F6F8}4MB}                                                    & \multicolumn{1}{c|}{\cellcolor[HTML]{5B98CF}0.28}                       & \multicolumn{1}{c|}{\cellcolor[HTML]{5E9ACF}0.30}                        & \multicolumn{1}{c|}{\cellcolor[HTML]{5A97CE}0.27}                        & \multicolumn{1}{c|}{\cellcolor[HTML]{7FAED9}0.48}                        & \multicolumn{1}{c|}{\cellcolor[HTML]{DAE7F4}0.99}                        & \multicolumn{1}{c|}{\cellcolor[HTML]{FFFFFF}1.36}                         & \multicolumn{1}{c|}{\cellcolor[HTML]{FFFCFC}4.83}                         & \cellcolor[HTML]{FFFFFF}1.22   \\ \cline{2-10} 
\multicolumn{1}{|c|}{\cellcolor[HTML]{F4F6F8}}                             & \multicolumn{1}{c|}{\cellcolor[HTML]{F4F6F8}32MB}                                                   & \multicolumn{1}{c|}{\cellcolor[HTML]{5B98CE}0.28}                       & \multicolumn{1}{c|}{\cellcolor[HTML]{5E9ACF}0.29}                        & \multicolumn{1}{c|}{\cellcolor[HTML]{5A97CE}0.27}                        & \multicolumn{1}{c|}{\cellcolor[HTML]{79AAD7}0.44}                        & \multicolumn{1}{c|}{\cellcolor[HTML]{DCE9F4}1.01}                        & \multicolumn{1}{c|}{\cellcolor[HTML]{FFFFFF}1.38}                         & \multicolumn{1}{c|}{\cellcolor[HTML]{FFFCFC}4.88}                         & \cellcolor[HTML]{FFFFFF}1.22   \\ \cline{2-10} 
\multicolumn{1}{|c|}{\cellcolor[HTML]{F4F6F8}}                             & \multicolumn{1}{c|}{\cellcolor[HTML]{F4F6F8}64MB}                                                   & \multicolumn{1}{c|}{\cellcolor[HTML]{5B98CF}0.28}                       & \multicolumn{1}{c|}{\cellcolor[HTML]{5E9ACF}0.29}                        & \multicolumn{1}{c|}{\cellcolor[HTML]{5A97CE}0.27}                        & \multicolumn{1}{c|}{\cellcolor[HTML]{79AAD7}0.44}                        & \multicolumn{1}{c|}{\cellcolor[HTML]{DBE8F4}1.00}                        & \multicolumn{1}{c|}{\cellcolor[HTML]{FFFFFF}1.40}                         & \multicolumn{1}{c|}{\cellcolor[HTML]{FFFCFC}4.93}                         & \cellcolor[HTML]{FFFFFF}1.23   \\ \cline{2-10} 
\multicolumn{1}{|c|}{\cellcolor[HTML]{F4F6F8}}                             & \multicolumn{1}{c|}{\cellcolor[HTML]{F4F6F8}128MB}                                                  & \multicolumn{1}{c|}{\cellcolor[HTML]{5F9AD0}0.30}                       & \multicolumn{1}{c|}{\cellcolor[HTML]{5A97CE}0.27}                        & \multicolumn{1}{c|}{\cellcolor[HTML]{5A97CE}0.27}                        & \multicolumn{1}{c|}{\cellcolor[HTML]{79AAD7}0.45}                        & \multicolumn{1}{c|}{\cellcolor[HTML]{DDEAF5}1.01}                        & \multicolumn{1}{c|}{\cellcolor[HTML]{FFFFFF}1.39}                         & \multicolumn{1}{c|}{\cellcolor[HTML]{FFFCFC}4.91}                         & \cellcolor[HTML]{FFFFFF}1.23   \\ \cline{2-10} 
\multicolumn{1}{|c|}{\cellcolor[HTML]{F4F6F8}}                             & \multicolumn{1}{c|}{\cellcolor[HTML]{F4F6F8}256MB}                                                  & \multicolumn{1}{c|}{\cellcolor[HTML]{5F9AD0}0.30}                       & \multicolumn{1}{c|}{\cellcolor[HTML]{5A97CE}0.27}                        & \multicolumn{1}{c|}{\cellcolor[HTML]{5A97CE}0.27}                        & \multicolumn{1}{c|}{\cellcolor[HTML]{79AAD7}0.44}                        & \multicolumn{1}{c|}{\cellcolor[HTML]{DDEAF5}1.01}                        & \multicolumn{1}{c|}{\cellcolor[HTML]{FFFFFF}1.40}                         & \multicolumn{1}{c|}{\cellcolor[HTML]{FFFCFC}4.90}                         & \cellcolor[HTML]{FFFFFF}1.23   \\ \cline{2-10} 
\multicolumn{1}{|c|}{\multirow{-6}{*}{\cellcolor[HTML]{F4F6F8}ROAST}}      & \multicolumn{1}{c|}{\cellcolor[HTML]{F4F6F8}512MB}                                                  & \multicolumn{1}{c|}{\cellcolor[HTML]{5F9AD0}0.30}                       & \multicolumn{1}{c|}{\cellcolor[HTML]{5E9ACF}0.30}                        & \multicolumn{1}{c|}{\cellcolor[HTML]{5A97CE}0.27}                        & \multicolumn{1}{c|}{\cellcolor[HTML]{79ABD7}0.45}                        & \multicolumn{1}{c|}{\cellcolor[HTML]{DFEBF5}1.02}                        & \multicolumn{1}{c|}{\cellcolor[HTML]{FFFFFF}1.39}                         & \multicolumn{1}{c|}{\cellcolor[HTML]{FFFCFC}4.95}                         & \cellcolor[HTML]{FFFFFF}1.24   \\ \hline
\end{tabular}}

\label{tab:fwd}
\end{table}
The high-performance community has heavily investigated the fast implementation of the General Matrix Multiplication (GEMM) kernel, a fundamental operation in many computational workloads, including deep learning. Optimized implementations of GEMM kernels are available in vendor libraries such as cuBLAS~\cite{cublas} and CUTLASS~\cite{cutlass}. Unfortunately, these implementations do not support custom tile loading operations, which is the key of \roach-MM. To implement \roach-MM to a level of efficiency comparable to that of optimized GEMM kernels, we used Triton~\cite{triton}: an intermediate language for tiled neural network computations.
Triton abstracts out the shared memory management to make it helpful in customizing tiled operations with high efficiency.

In our implementation of \roach-MM, the optimal size of coalesced tiles is a parameter that depends on the shape of the weight matrix. Therefore, different tile sizes can lead to different parallelism, occupancy, and shared memory efficiency, resulting in different execution times. 
We autotune this parameter to obtain the best performance for particular matrix shapes. We propose two strategies for autotuning each \roast-MM layer - (1) Optimize the inference workload by autotuning the forward kernel and sharing the tile size with the backward kernels. (2) Optimize the training workload by autotuning the forward and backward kernels together.
Table~\ref{tab:fwd} shows the inference performance of a simple model using \roast-MM for matrix multiplication on compressed memory.
Our model linearly transforms the input vector and computes its norm.
We optimized the \roast-MM kernel for this experiment using the inference-optimal strategy. 

We make the following observations from Table~\ref{tab:fwd}: 
(1) \roast-MM outperforms HashedNet kernel consistently across the different multiplication workloads. On an average over different workloads, \roast-MM is up to 45$\times$ faster than HashedNet and only $1.34\times$ slower than PyTorch-MM.
(2) \roast-MM's performance scales better than PyTorch-MM with the increase in workload.
As the workload increases 1600$\times$ (from 512$\times$512 to 20480$\times$20480), PyTorch-MM takes $39\times$ time, HashedNet takes 106$\times$ time whereas {\roast}-MM only takes around $16\times$ time. 

We present the detailed numbers for the training workload of a simple model in appendix \ref{sec:training} where we optimized \roast-MM using the training-optimal strategy. Note that training time composes of forward, backward, and optimization (alternatively, weight update function) times.
We make the following observations: (1) While the trends for backward function times are similar to forward function times, optimization for HashedNet and \roast-MM is faster than PyTorch-MM on large workloads due to the small sizes of compressed memory. (2) On average, \roast-MM has close performance compared to PyTorch for the small compressed memory and can be up to $2\times$ slower when the compressed memory is large.

\ssection{Application: Compression of BERT for text-classification} \label{sec:robez_exp}
\begin{table}[]
\caption{Best known results in NLP according to the survey \cite{gupta2022compression} in various paradigms. $(*)$ Knowledge distillation results exist for BERT in literature for compression ranging from 1.6$\times$-370$\times$. However, most of the results give lower-quality models. Hence, we show the results that have similar accuracy to BERT-base.
$(+)$ As BERT uses word-piece embedding, we exclude results that perform compression on only embedding tables.}
\resizebox{\linewidth}{!}{
\begin{tabular}{|c|c|c|c|c|}
\hline
Type                                              & Methods                                                           & Compression                             & Quality                               & Task                                                                                                                            \\ \hline
Pruning                                           & \begin{tabular}[c]{@{}c@{}}RPP/ \\ Iterative magnitude\cite{guo2019reweighted}\end{tabular} & up to  2.5 $\times$                      & better/simiar                         & \begin{tabular}[c]{@{}c@{}}Textual entailment\\Text classification\\ Question answering\\Reading comprehension\end{tabular} \\ \hline
Quantization                                      & \begin{tabular}[c]{@{}c@{}}QBERT/ \\ Mixed precision\cite{shen2020q}\end{tabular}  & up to  10 $\times$                       & worse                                 & \begin{tabular}[c]{@{}c@{}}Question answering\\ Textual entailment\\ Named entity recognition\end{tabular}                      \\ \hline
{\color[HTML]{333333} Knowledge distillation$*$} & {\color[HTML]{333333} BERT distillations\cite{bertpatient, bertsqueeze, sanh2019distilbert}}                         & {\color[HTML]{333333} up to $1.6\times$} & {\color[HTML]{333333} better/similar} & Glue benchmark                                                                                                                  \\ \hline
{\color[HTML]{333333} Parameter sharing$+$}      & {\color[HTML]{333333} ALBERT\cite{lan2019albert}}                                     & {\color[HTML]{333333} up to $18\times$}  & {\color[HTML]{333333} worse}          & Glue benchmark                                                                                                                  \\ \hline
\end{tabular}}

\label{tab:sota}
\end{table}
We want to demonstrate that generic model compression with {\roast} is a good approach in SOTA architectures for essential domains like NLP. However, SOTA results in NLP are achieved by comprehensive and costly pre-training followed by task-specific fine-tuning.
The cost of pre-training is beyond the scope of even many companies in the industry, let alone academic institutions. The estimated cost of a single training run of BERT Large (340M parameters) is estimated to be $10K$ USD and, along with required hyperparameter tuning, can go as large as $200K$ USD \cite{sharir2020cost}. In light of these exorbitant costs, we propose the following evaluation for {\roast}. We train {\roast}ed BERT models from scratch on the text-classification task using five different datasets of varying sizes. We show that even at high compression of 100$\times$-1000$\times$, {\roast}ed BERT models maintain the accuracy compared with the original BERT model.

\textbf{Baselines:} We present the known results on BERT compression with different paradigms in table \ref{tab:sota}. We can see that previous efforts have shown maximum compression of around $2.5\times$ on BERT models while maintaining the quality of the model. Results beyond this compression lead to deterioration of model quality. While these results are not comparable to ours in an apples-to-apples fashion, it is exciting that we are first to demonstrate compression of the order of 100$\times$-1000$\times$ on BERT models. 


\begin{SCtable}[]
\caption{Datasets used for text-classification task. All datasets are taken from HuggingFace\cite{lhoest2021datasets} (available with Apache-2.0 license) and use the standard test/train split provided by them. *For amazon-polarity, we used the first 100K test samples for evaluation}
\label{tab:nlpds}
\centering
\begin{tabular}{|c|c|c|}
\hline
\textbf{Datasets}         & \textbf{Train size} & \textbf{Test size} \\ \hline
tweet-eval (hate-speech) & 9K                  & 1K                 \\ \hline
tweet-eval (sentiment)   & 45K                 & 2K                 \\ \hline
ag-news (news)          & 120K                & 7.6K               \\ \hline
yelp-polarity (review)   & 560K                & 38K                \\ \hline
amazon-polarity (review)  & 3.6M                & 400K*               \\ \hline
\end{tabular}
\end{SCtable}

\begin{figure}[ht]
    \centering
    \begin{subfigure}[t]{0.32\textwidth}
    \includegraphics[scale=0.31]{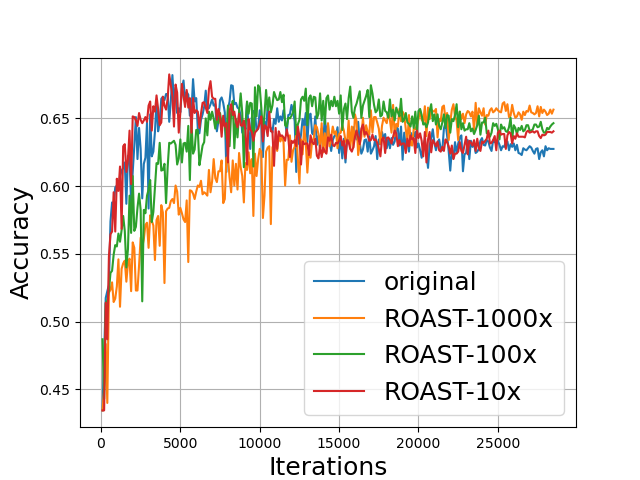}
    \caption{tweet-eval-sentiment (test)}
    \centering
    \end{subfigure}
    \begin{subfigure}[t]{0.32\textwidth}
    \includegraphics[scale=0.31]{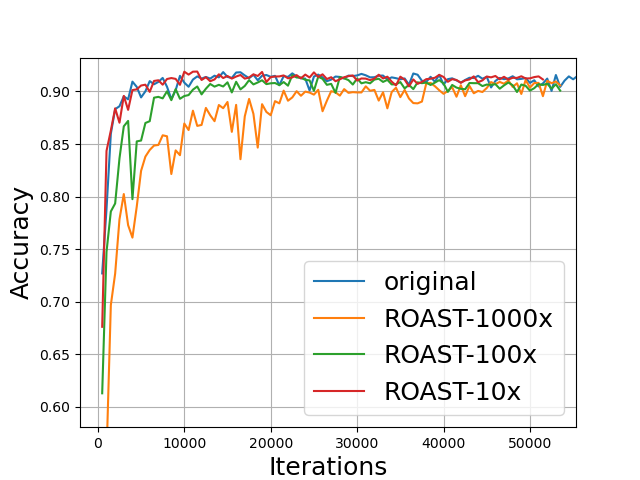}
    \caption{ag-news (test)}
    \end{subfigure}
    \centering
    \begin{subfigure}[t]{0.32\textwidth}
\includegraphics[scale=0.31]{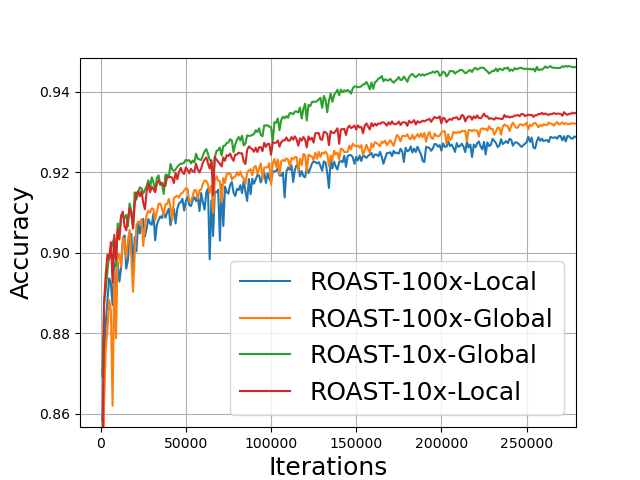}
    \caption{GMS vs. LMS (amazon-polarity)}
    \end{subfigure}
    \centering
    \begin{subfigure}[t]{0.32\textwidth}
    \includegraphics[scale=0.31]{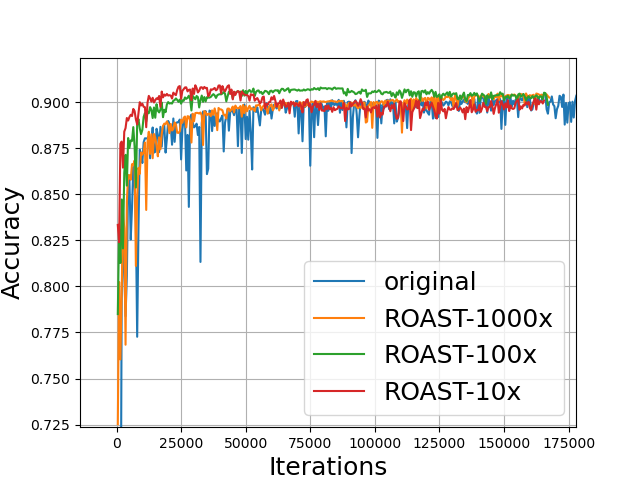}
    \caption{yelp-polarity (test)}
    \end{subfigure}
    \centering
    \begin{subfigure}[t]{0.32\textwidth}
    \includegraphics[scale=0.31]{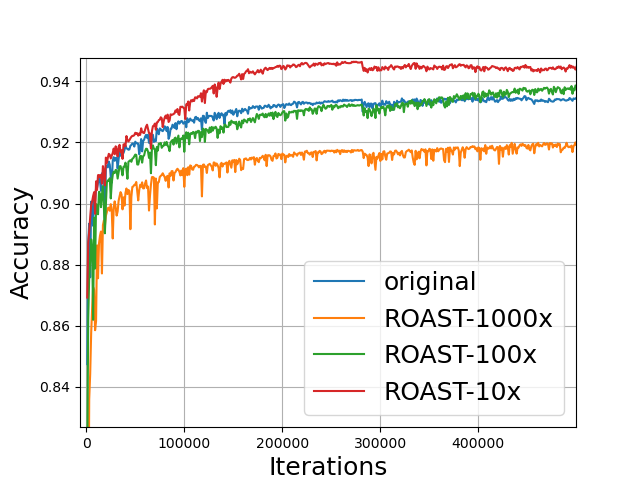}
    \caption{amazon-polarity (test)}
    \end{subfigure}
     \begin{subfigure}[t]{0.32\textwidth}
    \includegraphics[scale=0.31]{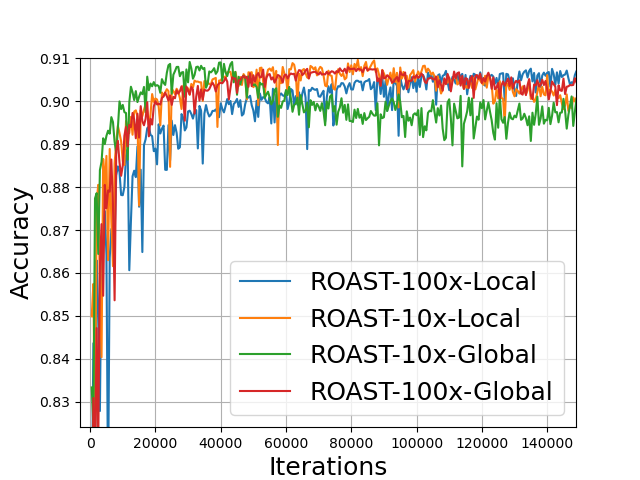}
    \caption{GMS vs. LMS (yelp-polarity)} \label{fig:globallocal}
    \end{subfigure}
    \caption{Figures a, b, d, and e compare the test accuracy of {\roast}ed BERT models and the original BERT model. Figures c and f compare LMS and GMS in {\roast}. The tweet-eval (hate) dataset is omitted here for lack of space. The variance in accuracy over different runs is within 0.001 in all cases. For example, three runs of BERT on the yelp dataset has a standard deviation of $0.00093$ and that of {\roast}ed BERT-100$\times$ is $0.00035$. More details can be found in Appendix~\ref{sec:variance}}
    \label{fig:nlpacc}
\end{figure}

 \textbf{Experimental setting}
We chose text-classification datasets (shown in table \ref{tab:nlpds}) from the huggingface dataset repository \cite{lhoest2021datasets}. ``amazon-polarity" is the largest text-classification dataset in huggingface with 3.6M samples.
We used the BERT-base (108M parameters) model that has ${\sim}$85M matrix multiplication parameters (78.9\%), ${\sim}$22M embedding parameters with a vocab of 30K (20.9\%), and 120K other parameters like bias.
We used a learning rate of 2e-5, a batch size of 64 and an input sequence length of 128.
All experiments were performed on a NVIDIA V100 GPU.
We used the BERT implementation from huggingface~\cite{huggingfacetransformers}.
 
\textbf{Results} The results of various {\roach } compression rates are shown in Figure~\ref{fig:nlpacc}. We make the following observations:
\begin{itemize}[leftmargin=*,nosep]
    \item In all datasets, \roach-10$\times$ and \roach-100$\times$ BERT reach similar or better accuracy than the original BERT. The better accuracy can be potentially attributed to the implicit regularization due to weight sharing.
    \item \roach-1000$\times$ BERT also reaches similar accuracy in three out of five datasets (tweet-eval (hate), yelp-polarity, ag-news).
    At times, \roach-1000$\times$ BERT's convergence is slow.
    Since \roach-1000$\times$ BERT does not overfit, it is possible that it may converge to similar accuracy as BERT with more iterations.
    \item The effect of global weight sharing in {\roach} can be demonstrated in Figure~\ref{fig:nlpacc} (c,f). This validates our theory that global weight sharing is superior than local weight sharing. 
\end{itemize}

\ssection{Conclusion}
This paper introduces a model-agnostic efficient model compression technique with {\roast} hashing. The results for {\roast} compression on the widely used BERT model are exciting, indicating that transformer architectures can indeed be compressed to as large as 100$\times$-1000$\times$ without loss in quality. 

\ssection{Negative Societal Impact} {\roast} promotes fast and efficient AI. Current and future Effects of AI is an active research area. We remain cautiously optimistic about future of AI.
\bibliography{neurips.bib}
\bibliographystyle{unsrt}

\appendix
\section{Theory}
{\roast} is a generalized model compression which performs operation specific system-friendly lookup and global memory sharing. This raises some interesting theoretical questions

\subsection{Backward pass for model sharing weights across different components} \label{sec:gradexp}
A general function sharing a weight, say $x$ across different components can be written as , $f(x, g(x))$ The interpretation is that x was used in g(.) and then again used ahead in f. (In case of MLP, we can think of x being used in multiple layers)

Let $f(g_1, g_2)$ where both $g_1$ and $g_2$ are functions of $x$. 

\begin{align}
    \frac{\partial f(g_1, g_2)}{\partial x} = \frac{\partial f(g_1, g_2)}{\partial g_1} * \frac{\partial g_1}{\partial x} +  \frac{\partial f(g_1, g_2)}{\partial g_2}  * \frac{\partial g_2}{\partial x}
\end{align}

$g_1 = x$ and $g_2 = g(x)$

\begin{align}
    \frac{\partial f(g_1, g_2)}{\partial x} = \frac{\partial f(x, g(y))}{\partial x} |_{y=x} +  \frac{\partial f(y, g(x))}{\partial g(x)}  * \frac{\partial g(x)}{\partial x}|_{y=x}
\end{align}

\begin{align}
    \frac{\partial f(g_1, g_2)}{\partial x} = \frac{\partial f(x, g(y))}{\partial x}|_{y=x} +  \frac{\partial f(y, g(x))}{\partial x} |_{y=x}
\end{align}
Renaming,
\begin{align}
    \frac{\partial f(x, g(x))}{\partial x} = \frac{\partial f(z, g(y))}{\partial z}|_{y=x, z=x} +  \frac{\partial f(z, g(y))}{\partial y} |_{y=x, z=x}
\end{align}
Thus, we can essentially consider each place where x appears as new variables and then gradient w.r.t x is just summation of partial derivatives of the function w.r.t these new variables. Thus, it is easy to implement this in the backward pass. In order to make sure that the memory utilization in backward pass is not of the order of the recovered model size, we do not use the auto-differentiation of tensorflow/pytorch. We implement our own backward pass and it can be found in the code.

\subsection{Global feature hashing vs local feature hashing.} \label{sec:theory_appx}
We can consider model compression techniques as dimensionality reduction of the parameter vector (a one dimensional vector of all parameters in a model) of size n into a vector of size $|\mathcal{M}|=m$. Quality of inner-product preservation is used as a metric to measure the quality of dimensionality reduction. In terms of dimensionality reduction, {\roach} uses ROBE hashing \cite{robez}, which showed that chunk based hashing is theoretically better than hashing individual elements. In this section, we analyse GMS proposal of {\roach} against LMS of HashedNet. For the purpose of this comparison we assume a chunk size of 1. Consider two parameter vectors $x, y \in R^n$. We are interested in how inner product between these parameter vectors are preserved under hashing. Let $x = [x_1 x_2 ... x_k]$ and $y = [y_1 y_2 ... y_k]$ be composed of k pieces of sizes $n_1, n_2,...n_k$. In LMS, let each piece be mapped into memory of size $f_i m$ where $\sum_{i} f_i = 1$. 

The estimators of inner product in the GMS case can be written as ,

\begin{equation}
    \widehat{\ip{x}{y}}_{G,m} = \sum_{j=1}^m (\sum_{i=1}^n \mathbb{I}(h(i){=}j) g(i) x[i]) (\sum_{i=1}^n \mathbb{I}(h(i){=}j) g(i) y[i])
\end{equation}
The estimate of inner product with LMS can be written as,
\begin{equation}
    \widehat{\ip{x}{y}}_{L,m,\Vec{f}} = \sum_{l=1}^k \sum_{j=1}^{f_l m} (\sum_{i=1}^{n_l} \mathbb{I}(h(i){=}j) g(i) x_l[i]) (\sum_{j=1}^{n_l} \mathbb{I}(h(i){=}j) g(i) y_l[i]) = \sum_{l=1}^k \widehat{\ip{x_l}{y_l}}_{G,(f_i m)}
\end{equation}
Note that 
\begin{equation}
    \widehat{\ip{x}{y}}_{L, m, \Vec{f}} = \sum_{l=1}^k \widehat{\ip{x_l}{y_l}}_{G,(f_l m)}
\end{equation}
The GMS estimator is the standard feature hashing estimator and the LMS is essentially sum of GMS estimators for each of the piece. as $E[g(i)] = 0$, it is easy to check by linearity of expectations that
 \textbf{Expectation}
 The suffix L refers to local hashing and G refers to global hashing.
 
 \begin{align}
     E_G = \mathbb{E}(\widehat{\ip{x}{y}}_{G,m}) =  \ip{x}{y}
 \end{align}
  
  \begin{align}
     E_L = \mathbb{E}(\widehat{\ip{x}{y}}_{L,m,\Vec{f}}) =  \ip{x}{y}
 \end{align}
 
 Let us now look at the variance. Let us follow the following notation,
 \begin{itemize}
     \item $V_G = \mathbb{V}(\widehat{\ip{x}{y}}_{G,m})$. GMS variance of entire vectors
     \item $V_L = \mathbb{V}(\widehat{\ip{x}{y}}_{L,m,\Vec{f}})$. LMS variance of entire vectors
     \item $V_l = \mathbb{V}(\widehat{\ip{x_l}{y_l}}_{G,f_l m})$. variance of each piece
 \end{itemize}
 we can write $V_l$ as follows. The following equation is easy to derive and it can be found the lemma 2 of \cite{featurehash} 

\begin{equation}
    V_l = \frac{1}{f_l} \frac{1}{m} (\sum_{i\neq j} a_i^2 b_j^2  + \sum_{i \neq j} a_i b_i a_j b_j) \textrm{ where } x_l = (a_1, a_2 ... a_{n_l}) \textrm{ and } y_l = (b_1, b_2 ... b_{n_l})
\end{equation}
 As, each of the piece is independently hashed in LSM, we can see
 \begin{equation}
     V_L = \sum_{l=1}^k V_l
 \end{equation}
 Let us now look at $V_G$. Again, using lemma 2 from \cite{featurehash}
 \begin{equation}
     V_G = \frac{1}{m} (\sum_{i\neq j} x_i^2 y_j^2  + \sum_{i \neq j} x_i y_i x_j y_j)
 \end{equation}
 The expression can be split into terms that belong to same pieces and those across pieces
  \begin{align*}
     & V_G = \frac{1}{m} \sum_{l=1}^k  (\sum_{i\neq j \in \textrm{piece-l}} x_i^2 y_j^2  + \sum_{i \neq j \in \textrm{piece-l}} x_i y_i x_j y_j) \\
     & + \frac{1}{m} \sum_{l1=1}^k \sum_{l2=1, l1\neq l2}^k ( \sum_{i \in \textrm{piece-l1}, j \in \textrm{pieces-l2}} (x_{i}^2 y_{j}^2) + \sum_{i \in \textrm{piece-l1}, j \in \textrm{pieces-l2}} x_{i}y_{i}x_{j}y_{j}))
 \end{align*}
  
 \begin{equation}
     V_G = \sum_{l=1}^k f_l V_l + \frac{1}{m} \sum_{l1=1}^l \sum_{l2=1, l1\neq l2}^l ||x_{l1}||_2^2 ||y_{l2}||_2^2 + \ip{x_{l1}}{y_{l2}} \ip{x_{l2}}{y_{l2}}
 \end{equation}
 
\textbf{Observation 1:} In $V_L$ we can see that there are terms with $\frac{1}{f_l}$ which makes it unbounded. It makes sense as if number of pieces increase a lot a lot of compressions will not work for example if number of peices $ > |\mathcal{M}|$. Also, it will affect $V_L$ a lot when some $f_l$ is very small which can often be the case. For example, generally embedding tables in DLRM model are much larger than that of matrix multiplciation modules (MLP) . which can make $f\approx 0.001$ for MLP components. 

\textbf{Observation 2:} Practically we can assume each piece, no matter the size of the vector, to be of same norm. The reason lies in initialization. According to Xavier's initialization the weights of a particular node are initialized with norm 1. So for now lets assume a more practical case of all norms being equal to $\sqrt{\alpha}$. Also, in order to make the comparisons we need to consider some average case over the data. So let us assume that under independent randomized data assumption, the expected value of all inner products are $\beta$. 
With this , in expectation over randomized data, we have
\begin{equation}
    V_G = \sum f_l V_l + \frac{k(k-1)}{m} (\alpha^2 + \beta^2)
\end{equation}

Now note that, 
\begin{equation}
    V_l =  \frac{1}{f_l} \frac{1}{m} (\sum_{i\neq j} a_i^2 b_j^2  + \sum_{i \neq j} a_i b_i a_j b_j) \textrm{ where } x_l = (a_1, a_2 ... a_{n_l}) \textrm{ and } y_l = (b_1, b_2 ... b_{n_l})
\end{equation}

(dropping the subscript "l" below)
\begin{equation}
    V_l =  \frac{1}{f_l} \frac{1}{m} ((||x||_2^2 ||y||_2^2 + \ip{x}{y}^2) - 2 \sum_{i} x_i^2 y_i^2 )
\end{equation}

\begin{equation}
    V_l =  \frac{1}{f_l} \frac{1}{m} ((\alpha^2 + \beta^2) - 2 \sum_{i} x_i^2 y_i^2 )
\end{equation}
Note that for each negative term, there are $n_l$ positive terms. To simplify we disregard this term in the equation above. This is an approximation which is practical and only made to get a sense of $V_L$ and $V_G$ relation.

\begin{align*}
    & V_L - V_G = \sum V_l - \sum f_l V_l - \frac{k(k-1)}{m} (\alpha^2 + \beta^2) \\
    & V_L - V_G = \sum_l \frac{1}{m} (\frac{1}{f_l} - 1) ((\alpha^2 + \beta^2) ) - \frac{k(k-1)}{m} (\alpha^2 + \beta^2) \\
    & V_L - V_G = \sum_l \frac{1}{m}(\frac{1}{f_l} - 1) ((\alpha^2 + \beta^2)  - \frac{k(k-1)}{m} (\alpha^2 + \beta^2) \\
    & V_L - V_G \geq  \frac{k(k-1)}{m} ((\alpha^2 + \beta^2) - \frac{k(k-1)}{m} (\alpha^2 + \beta^2) \\
    & V_L - V_G \geq   0 \
\end{align*}
Note that we ignored a term which reduces the $V_L $ a bit, Let the error be $\epsilon$
\begin{equation}
    V_L - V_G \geq   -\epsilon
\end{equation}
The above equation shows even for the best case, $V_G$ might be slightly more than $V_L$. However for general case where harmonic mean is much worse than arithmetic mean, $V_L$ will be much larger depending on exact $f_l$ s


                

\newpage
\section{\roast-MM latency measurements} \label{sec:eff}
\subsection{Inference optimization}

\subsection{Training optimization}
See tables \ref{tab:total-fwd}, \ref{tab:total-bwd}, \ref{tab:total-opt}, \ref{tab:total-total}
\label{sec:training}
\begin{table}[h]
\begin{tabular}{|c|c|rrrrrrrr|}
\hline
\multicolumn{1}{|l|}{}                                                                 & \multicolumn{1}{l|}{}                                                            & \multicolumn{8}{c|}{\textbf{\begin{tabular}[c]{@{}c@{}}forward(ms) \\ (optimized for forward +  backward)\end{tabular}}}                                                                                                                                                                                                                                                                                                                                                                                                                                     \\ \hline
\rowcolor[HTML]{DFE4EC} 
\textit{}                                                                              &                                                                                  & \multicolumn{8}{c|}{\cellcolor[HTML]{DFE4EC}dim (Matrix dimension = dim x dim)}                                                                                                                                                                                                                                                                                                                                                                                                                                                                                             \\ \hline
\cellcolor[HTML]{DFE4EC}\textit{}                                                      & \cellcolor[HTML]{DFE4EC}\begin{tabular}[c]{@{}c@{}}Memory\\    (mb)\end{tabular} & \multicolumn{1}{c|}{\cellcolor[HTML]{8093B3}{\color[HTML]{FFFFFF} 512}} & \multicolumn{1}{r|}{\cellcolor[HTML]{8093B3}{\color[HTML]{FFFFFF} 1024}} & \multicolumn{1}{r|}{\cellcolor[HTML]{8093B3}{\color[HTML]{FFFFFF} 2048}} & \multicolumn{1}{r|}{\cellcolor[HTML]{8093B3}{\color[HTML]{FFFFFF} 4096}} & \multicolumn{1}{r|}{\cellcolor[HTML]{8093B3}{\color[HTML]{FFFFFF} 8096}} & \multicolumn{1}{r|}{\cellcolor[HTML]{8093B3}{\color[HTML]{FFFFFF} 10240}} & \multicolumn{1}{r|}{\cellcolor[HTML]{8093B3}{\color[HTML]{FFFFFF} 20480}} & \multicolumn{1}{l|}{Average}  \\ \hline
\cellcolor[HTML]{F4F6F8}\begin{tabular}[c]{@{}c@{}}Full \\ (uncompressed)\end{tabular} & \cellcolor[HTML]{F4F6F8}                                                         & \multicolumn{1}{r|}{\cellcolor[HTML]{20629D}0.16}                       & \multicolumn{1}{r|}{\cellcolor[HTML]{1A5D9A}0.12}                        & \multicolumn{1}{r|}{\cellcolor[HTML]{1A5D9A}0.12}                        & \multicolumn{1}{r|}{\cellcolor[HTML]{2E6CA3}0.24}                        & \multicolumn{1}{r|}{\cellcolor[HTML]{779FC3}0.66}                        & \multicolumn{1}{r|}{\cellcolor[HTML]{A3BED6}0.91}                         & \multicolumn{1}{r|}{\cellcolor[HTML]{FFFEFE}3.03}                         & \cellcolor[HTML]{86AACA}0.75  \\ \hline
\cellcolor[HTML]{F4F6F8}                                                               & \cellcolor[HTML]{F4F6F8}4                                                        & \multicolumn{1}{r|}{\cellcolor[HTML]{447BAD}0.37}                       & \multicolumn{1}{r|}{\cellcolor[HTML]{4279AC}0.35}                        & \multicolumn{1}{r|}{\cellcolor[HTML]{759EC2}0.65}                        & \multicolumn{1}{r|}{\cellcolor[HTML]{FFFFFF}2.04}                        & \multicolumn{1}{r|}{\cellcolor[HTML]{FFFBFB}6.23}                        & \multicolumn{1}{r|}{\cellcolor[HTML]{FFF8F8}9.62}                         & \multicolumn{1}{r|}{\cellcolor[HTML]{FFDFDF}35.64}                        & \cellcolor[HTML]{FFF9F9}7.84  \\ \cline{2-10} 
\cellcolor[HTML]{F4F6F8}                                                               & \cellcolor[HTML]{F4F6F8}32                                                       & \multicolumn{1}{r|}{\cellcolor[HTML]{487EAF}0.39}                       & \multicolumn{1}{r|}{\cellcolor[HTML]{4E82B1}0.42}                        & \multicolumn{1}{r|}{\cellcolor[HTML]{A0BCD5}0.90}                        & \multicolumn{1}{r|}{\cellcolor[HTML]{FFFDFD}3.67}                        & \multicolumn{1}{r|}{\cellcolor[HTML]{FFF4F4}13.73}                       & \multicolumn{1}{r|}{\cellcolor[HTML]{FFECEC}22.06}                        & \multicolumn{1}{r|}{\cellcolor[HTML]{FFAAAA}92.83}                        & \cellcolor[HTML]{FFEFEF}19.14 \\ \cline{2-10} 
\cellcolor[HTML]{F4F6F8}                                                               & \cellcolor[HTML]{F4F6F8}64                                                       & \multicolumn{1}{r|}{\cellcolor[HTML]{3E77AA}0.33}                       & \multicolumn{1}{r|}{\cellcolor[HTML]{5788B5}0.47}                        & \multicolumn{1}{r|}{\cellcolor[HTML]{C4D6E5}1.11}                        & \multicolumn{1}{r|}{\cellcolor[HTML]{FFFBFB}6.45}                        & \multicolumn{1}{r|}{\cellcolor[HTML]{FFE9E9}25.78}                       & \multicolumn{1}{r|}{\cellcolor[HTML]{FFD9D9}42.51}                        & \multicolumn{1}{r|}{\cellcolor[HTML]{FF5959}178.20}                       & \cellcolor[HTML]{FFDFDF}36.41 \\ \cline{2-10} 
\cellcolor[HTML]{F4F6F8}                                                               & \cellcolor[HTML]{F4F6F8}128                                                      & \multicolumn{1}{r|}{\cellcolor[HTML]{3671A7}0.28}                       & \multicolumn{1}{r|}{\cellcolor[HTML]{6693BC}0.56}                        & \multicolumn{1}{r|}{\cellcolor[HTML]{FFFFFF}1.61}                        & \multicolumn{1}{r|}{\cellcolor[HTML]{FFF8F8}9.07}                        & \multicolumn{1}{r|}{\cellcolor[HTML]{FFE1E1}34.21}                       & \multicolumn{1}{r|}{\cellcolor[HTML]{FFCCCC}56.07}                        & \multicolumn{1}{r|}{\cellcolor[HTML]{FF2929}229.34}                       & \cellcolor[HTML]{FFD4D4}47.31 \\ \cline{2-10} 
\cellcolor[HTML]{F4F6F8}                                                               & \cellcolor[HTML]{F4F6F8}256                                                      & \multicolumn{1}{r|}{\cellcolor[HTML]{2767A0}0.20}                       & \multicolumn{1}{r|}{\cellcolor[HTML]{6290BA}0.54}                        & \multicolumn{1}{r|}{\cellcolor[HTML]{FFFFFF}1.72}                        & \multicolumn{1}{r|}{\cellcolor[HTML]{FFF7F7}9.95}                        & \multicolumn{1}{r|}{\cellcolor[HTML]{FFDDDD}38.17}                       & \multicolumn{1}{r|}{\cellcolor[HTML]{FFC6C6}62.47}                        & \multicolumn{1}{r|}{\cellcolor[HTML]{FF0E0E}258.11}                       & \cellcolor[HTML]{FFCFCF}53.02 \\ \cline{2-10} 
\multirow{-6}{*}{\cellcolor[HTML]{F4F6F8}HashedNet}                                    & \cellcolor[HTML]{F4F6F8}512                                                      & \multicolumn{1}{r|}{\cellcolor[HTML]{1D609C}0.14}                       & \multicolumn{1}{r|}{\cellcolor[HTML]{5B8BB7}0.50}                        & \multicolumn{1}{r|}{\cellcolor[HTML]{FFFFFF}1.88}                        & \multicolumn{1}{r|}{\cellcolor[HTML]{FFF7F7}10.37}                       & \multicolumn{1}{r|}{\cellcolor[HTML]{FFDBDB}40.40}                       & \multicolumn{1}{r|}{\cellcolor[HTML]{FFC3C3}65.43}                        & \multicolumn{1}{r|}{\cellcolor[HTML]{FF0101}272.19}                       & \cellcolor[HTML]{FFCCCC}55.84 \\ \hline
\cellcolor[HTML]{F4F6F8}                                                               & \cellcolor[HTML]{F4F6F8}4                                                        & \multicolumn{1}{r|}{\cellcolor[HTML]{3974A8}0.30}                       & \multicolumn{1}{r|}{\cellcolor[HTML]{3A74A8}0.31}                        & \multicolumn{1}{r|}{\cellcolor[HTML]{3A74A8}0.31}                        & \multicolumn{1}{r|}{\cellcolor[HTML]{5B8BB7}0.50}                        & \multicolumn{1}{r|}{\cellcolor[HTML]{FDFDFE}1.43}                        & \multicolumn{1}{r|}{\cellcolor[HTML]{FFFFFF}2.01}                         & \multicolumn{1}{r|}{\cellcolor[HTML]{FFFAFA}7.54}                         & \cellcolor[HTML]{FFFFFF}1.77  \\ \cline{2-10} 
\cellcolor[HTML]{F4F6F8}                                                               & \cellcolor[HTML]{F4F6F8}32                                                       & \multicolumn{1}{r|}{\cellcolor[HTML]{3973A8}0.30}                       & \multicolumn{1}{r|}{\cellcolor[HTML]{3E77AA}0.33}                        & \multicolumn{1}{r|}{\cellcolor[HTML]{4179AB}0.35}                        & \multicolumn{1}{r|}{\cellcolor[HTML]{6492BB}0.55}                        & \multicolumn{1}{r|}{\cellcolor[HTML]{FFFFFF}1.44}                        & \multicolumn{1}{r|}{\cellcolor[HTML]{FFFFFF}2.09}                         & \multicolumn{1}{r|}{\cellcolor[HTML]{FFFAFA}7.59}                         & \cellcolor[HTML]{FFFFFF}1.81  \\ \cline{2-10} 
\cellcolor[HTML]{F4F6F8}                                                               & \cellcolor[HTML]{F4F6F8}64                                                       & \multicolumn{1}{r|}{\cellcolor[HTML]{3772A7}0.29}                       & \multicolumn{1}{r|}{\cellcolor[HTML]{3A74A8}0.31}                        & \multicolumn{1}{r|}{\cellcolor[HTML]{3E77AA}0.33}                        & \multicolumn{1}{r|}{\cellcolor[HTML]{6693BC}0.56}                        & \multicolumn{1}{r|}{\cellcolor[HTML]{FFFFFF}1.45}                        & \multicolumn{1}{r|}{\cellcolor[HTML]{FFFFFF}2.08}                         & \multicolumn{1}{r|}{\cellcolor[HTML]{FFFAFA}7.80}                         & \cellcolor[HTML]{FFFFFF}1.83  \\ \cline{2-10} 
\cellcolor[HTML]{F4F6F8}                                                               & \cellcolor[HTML]{F4F6F8}128                                                      & \multicolumn{1}{r|}{\cellcolor[HTML]{306DA4}0.25}                       & \multicolumn{1}{r|}{\cellcolor[HTML]{3470A6}0.27}                        & \multicolumn{1}{r|}{\cellcolor[HTML]{3571A6}0.28}                        & \multicolumn{1}{r|}{\cellcolor[HTML]{6290BA}0.54}                        & \multicolumn{1}{r|}{\cellcolor[HTML]{F9FBFC}1.41}                        & \multicolumn{1}{r|}{\cellcolor[HTML]{FFFFFF}2.09}                         & \multicolumn{1}{r|}{\cellcolor[HTML]{FFF9F9}7.84}                         & \cellcolor[HTML]{FFFFFF}1.81  \\ \cline{2-10} 
\cellcolor[HTML]{F4F6F8}                                                               & \cellcolor[HTML]{F4F6F8}256                                                      & \multicolumn{1}{r|}{\cellcolor[HTML]{21629D}0.16}                       & \multicolumn{1}{r|}{\cellcolor[HTML]{25659F}0.18}                        & \multicolumn{1}{r|}{\cellcolor[HTML]{25659F}0.19}                        & \multicolumn{1}{r|}{\cellcolor[HTML]{5486B4}0.46}                        & \multicolumn{1}{r|}{\cellcolor[HTML]{EBF1F6}1.33}                        & \multicolumn{1}{r|}{\cellcolor[HTML]{FFFFFF}2.02}                         & \multicolumn{1}{r|}{\cellcolor[HTML]{FFFAFA}7.82}                         & \cellcolor[HTML]{FFFFFF}1.74  \\ \cline{2-10} 
\multirow{-6}{*}{\cellcolor[HTML]{F4F6F8}ROAST}                                        & \cellcolor[HTML]{F4F6F8}512                                                      & \multicolumn{1}{r|}{\cellcolor[HTML]{2A68A1}0.21}                       & \multicolumn{1}{r|}{\cellcolor[HTML]{0E5595}0.06}                        & \multicolumn{1}{r|}{\cellcolor[HTML]{1C5F9B}0.13}                        & \multicolumn{1}{r|}{\cellcolor[HTML]{4B80B0}0.41}                        & \multicolumn{1}{r|}{\cellcolor[HTML]{E4ECF3}1.29}                        & \multicolumn{1}{r|}{\cellcolor[HTML]{FFFFFF}1.97}                         & \multicolumn{1}{r|}{\cellcolor[HTML]{FFFCFC}4.98}                         & \cellcolor[HTML]{E5ECF3}1.29  \\ \hline
\end{tabular}
\caption{Inference (forward pass time) for different shapes of square weight matrix with input batch of 512. The tile-parameters of multiplication are optimized for each function over "forward +  backward" pass .The measurements are taken with tf32 on A100 (48GB)}
\label{tab:total-fwd}
\end{table}
\begin{table}[h]
\begin{tabular}{|c|c|rrrrrrrr|}
\hline
\multicolumn{1}{|l|}{}                                                                 & \multicolumn{1}{l|}{}                                                            & \multicolumn{8}{c|}{\textbf{\begin{tabular}[c]{@{}c@{}}backward(ms) \\ (optimized for forward +  backward)\end{tabular}}}                                                                                                                                                                                                                                                                                                                                                                                                                                                   \\ \hline
\rowcolor[HTML]{DFE4EC} 
\textit{}                                                                              &                                                                                  & \multicolumn{8}{c|}{\cellcolor[HTML]{DFE4EC}dim (Matrix dimension = dim x dim)}                                                                                                                                                                                                                                                                                                                                                                                                                                                                                             \\ \hline
\cellcolor[HTML]{DFE4EC}\textit{}                                                      & \cellcolor[HTML]{DFE4EC}\begin{tabular}[c]{@{}c@{}}Memory\\    (mb)\end{tabular} & \multicolumn{1}{c|}{\cellcolor[HTML]{8093B3}{\color[HTML]{FFFFFF} 512}} & \multicolumn{1}{r|}{\cellcolor[HTML]{8093B3}{\color[HTML]{FFFFFF} 1024}} & \multicolumn{1}{r|}{\cellcolor[HTML]{8093B3}{\color[HTML]{FFFFFF} 2048}} & \multicolumn{1}{r|}{\cellcolor[HTML]{8093B3}{\color[HTML]{FFFFFF} 4096}} & \multicolumn{1}{r|}{\cellcolor[HTML]{8093B3}{\color[HTML]{FFFFFF} 8096}} & \multicolumn{1}{r|}{\cellcolor[HTML]{8093B3}{\color[HTML]{FFFFFF} 10240}} & \multicolumn{1}{r|}{\cellcolor[HTML]{8093B3}{\color[HTML]{FFFFFF} 20480}} & \multicolumn{1}{l|}{Average}  \\ \hline
\cellcolor[HTML]{F4F6F8}\begin{tabular}[c]{@{}c@{}}Full \\ (uncompressed)\end{tabular} & \cellcolor[HTML]{F4F6F8}                                                         & \multicolumn{1}{r|}{\cellcolor[HTML]{165A98}0.35}                       & \multicolumn{1}{r|}{\cellcolor[HTML]{0C5494}0.22}                        & \multicolumn{1}{r|}{\cellcolor[HTML]{0D5495}0.24}                        & \multicolumn{1}{r|}{\cellcolor[HTML]{20619D}0.48}                        & \multicolumn{1}{r|}{\cellcolor[HTML]{6391BA}1.35}                        & \multicolumn{1}{r|}{\cellcolor[HTML]{96B5D0}2.01}                         & \multicolumn{1}{r|}{\cellcolor[HTML]{FFFDFD}7.65}                         & \cellcolor[HTML]{82A7C8}1.76  \\ \hline
\cellcolor[HTML]{F4F6F8}                                                               & \cellcolor[HTML]{F4F6F8}4                                                        & \multicolumn{1}{r|}{\cellcolor[HTML]{2D6BA3}0.65}                       & \multicolumn{1}{r|}{\cellcolor[HTML]{24649E}0.53}                        & \multicolumn{1}{r|}{\cellcolor[HTML]{447BAD}0.95}                        & \multicolumn{1}{r|}{\cellcolor[HTML]{C3D5E4}2.60}                        & \multicolumn{1}{r|}{\cellcolor[HTML]{FFFCFC}8.51}                        & \multicolumn{1}{r|}{\cellcolor[HTML]{FFF9F9}13.21}                        & \multicolumn{1}{r|}{\cellcolor[HTML]{FFDDDD}56.59}                        & \cellcolor[HTML]{FFFAFA}11.86 \\ \cline{2-10} 
\cellcolor[HTML]{F4F6F8}                                                               & \cellcolor[HTML]{F4F6F8}32                                                       & \multicolumn{1}{r|}{\cellcolor[HTML]{2F6DA4}0.68}                       & \multicolumn{1}{r|}{\cellcolor[HTML]{306DA4}0.69}                        & \multicolumn{1}{r|}{\cellcolor[HTML]{86A9CA}1.80}                        & \multicolumn{1}{r|}{\cellcolor[HTML]{FFFEFE}6.36}                        & \multicolumn{1}{r|}{\cellcolor[HTML]{FFF2F2}24.13}                       & \multicolumn{1}{r|}{\cellcolor[HTML]{FFE8E8}38.95}                        & \multicolumn{1}{r|}{\cellcolor[HTML]{FF9898}160.54}                       & \cellcolor[HTML]{FFECEC}33.31 \\ \cline{2-10} 
\cellcolor[HTML]{F4F6F8}                                                               & \cellcolor[HTML]{F4F6F8}64                                                       & \multicolumn{1}{r|}{\cellcolor[HTML]{3470A6}0.74}                       & \multicolumn{1}{r|}{\cellcolor[HTML]{4C81B0}1.06}                        & \multicolumn{1}{r|}{\cellcolor[HTML]{D4E0EC}2.81}                        & \multicolumn{1}{r|}{\cellcolor[HTML]{FFFBFB}10.78}                       & \multicolumn{1}{r|}{\cellcolor[HTML]{FFE7E7}41.35}                       & \multicolumn{1}{r|}{\cellcolor[HTML]{FFD6D6}67.02}                        & \multicolumn{1}{r|}{\cellcolor[HTML]{FF4F4F}271.86}                       & \cellcolor[HTML]{FFDDDD}56.52 \\ \cline{2-10} 
\cellcolor[HTML]{F4F6F8}                                                               & \cellcolor[HTML]{F4F6F8}128                                                      & \multicolumn{1}{r|}{\cellcolor[HTML]{4179AB}0.91}                       & \multicolumn{1}{r|}{\cellcolor[HTML]{6290BA}1.34}                        & \multicolumn{1}{r|}{\cellcolor[HTML]{FFFFFF}3.40}                        & \multicolumn{1}{r|}{\cellcolor[HTML]{FFFAFA}12.41}                       & \multicolumn{1}{r|}{\cellcolor[HTML]{FFE0E0}51.00}                       & \multicolumn{1}{r|}{\cellcolor[HTML]{FFCCCC}81.25}                        & \multicolumn{1}{r|}{\cellcolor[HTML]{FF2424}337.31}                       & \cellcolor[HTML]{FFD4D4}69.66 \\ \cline{2-10} 
\cellcolor[HTML]{F4F6F8}                                                               & \cellcolor[HTML]{F4F6F8}256                                                      & \multicolumn{1}{r|}{\cellcolor[HTML]{5E8EB8}1.29}                       & \multicolumn{1}{r|}{\cellcolor[HTML]{89ABCB}1.84}                        & \multicolumn{1}{r|}{\cellcolor[HTML]{FFFFFF}4.02}                        & \multicolumn{1}{r|}{\cellcolor[HTML]{FFF8F8}14.57}                       & \multicolumn{1}{r|}{\cellcolor[HTML]{FFDCDC}58.03}                       & \multicolumn{1}{r|}{\cellcolor[HTML]{FFC6C6}91.18}                        & \multicolumn{1}{r|}{\cellcolor[HTML]{FF0A0A}376.83}                       & \cellcolor[HTML]{FFCECE}78.25 \\ \cline{2-10} 
\multirow{-6}{*}{\cellcolor[HTML]{F4F6F8}HashedNet}                                    & \cellcolor[HTML]{F4F6F8}512                                                      & \multicolumn{1}{r|}{\cellcolor[HTML]{9BB8D3}2.08}                       & \multicolumn{1}{r|}{\cellcolor[HTML]{C5D6E5}2.62}                        & \multicolumn{1}{r|}{\cellcolor[HTML]{FFFEFE}4.90}                        & \multicolumn{1}{r|}{\cellcolor[HTML]{FFF7F7}16.24}                       & \multicolumn{1}{r|}{\cellcolor[HTML]{FFD9D9}62.45}                       & \multicolumn{1}{r|}{\cellcolor[HTML]{FFC1C1}98.46}                        & \multicolumn{1}{r|}{\cellcolor[HTML]{FF0101}391.46}                       & \cellcolor[HTML]{FFCBCB}82.60 \\ \hline
\cellcolor[HTML]{F4F6F8}                                                               & \cellcolor[HTML]{F4F6F8}4                                                        & \multicolumn{1}{r|}{\cellcolor[HTML]{24659F}0.54}                       & \multicolumn{1}{r|}{\cellcolor[HTML]{24659F}0.54}                        & \multicolumn{1}{r|}{\cellcolor[HTML]{2968A1}0.60}                        & \multicolumn{1}{r|}{\cellcolor[HTML]{5789B5}1.20}                        & \multicolumn{1}{r|}{\cellcolor[HTML]{BFD1E2}2.54}                        & \multicolumn{1}{r|}{\cellcolor[HTML]{FFFFFF}3.72}                         & \multicolumn{1}{r|}{\cellcolor[HTML]{FFF9F9}13.99}                        & \cellcolor[HTML]{FAFBFC}3.30  \\ \cline{2-10} 
\cellcolor[HTML]{F4F6F8}                                                               & \cellcolor[HTML]{F4F6F8}32                                                       & \multicolumn{1}{r|}{\cellcolor[HTML]{2766A0}0.57}                       & \multicolumn{1}{r|}{\cellcolor[HTML]{2968A1}0.61}                        & \multicolumn{1}{r|}{\cellcolor[HTML]{306DA4}0.69}                        & \multicolumn{1}{r|}{\cellcolor[HTML]{4D81B1}1.06}                        & \multicolumn{1}{r|}{\cellcolor[HTML]{CCDBE8}2.71}                        & \multicolumn{1}{r|}{\cellcolor[HTML]{FFFFFF}4.04}                         & \multicolumn{1}{r|}{\cellcolor[HTML]{FFF8F8}15.07}                        & \cellcolor[HTML]{FFFFFF}3.54  \\ \cline{2-10} 
\cellcolor[HTML]{F4F6F8}                                                               & \cellcolor[HTML]{F4F6F8}64                                                       & \multicolumn{1}{r|}{\cellcolor[HTML]{2C6AA2}0.64}                       & \multicolumn{1}{r|}{\cellcolor[HTML]{336FA5}0.73}                        & \multicolumn{1}{r|}{\cellcolor[HTML]{3671A7}0.77}                        & \multicolumn{1}{r|}{\cellcolor[HTML]{5587B4}1.17}                        & \multicolumn{1}{r|}{\cellcolor[HTML]{D4E1EC}2.82}                        & \multicolumn{1}{r|}{\cellcolor[HTML]{FFFFFF}4.18}                         & \multicolumn{1}{r|}{\cellcolor[HTML]{FFF8F8}15.50}                        & \cellcolor[HTML]{FFFFFF}3.69  \\ \cline{2-10} 
\cellcolor[HTML]{F4F6F8}                                                               & \cellcolor[HTML]{F4F6F8}128                                                      & \multicolumn{1}{r|}{\cellcolor[HTML]{3872A7}0.79}                       & \multicolumn{1}{r|}{\cellcolor[HTML]{3973A8}0.81}                        & \multicolumn{1}{r|}{\cellcolor[HTML]{4078AB}0.89}                        & \multicolumn{1}{r|}{\cellcolor[HTML]{6592BB}1.38}                        & \multicolumn{1}{r|}{\cellcolor[HTML]{EFF4F8}3.17}                        & \multicolumn{1}{r|}{\cellcolor[HTML]{FFFFFF}4.73}                         & \multicolumn{1}{r|}{\cellcolor[HTML]{FFF6F6}18.30}                        & \cellcolor[HTML]{FFFFFF}4.30  \\ \cline{2-10} 
\cellcolor[HTML]{F4F6F8}                                                               & \cellcolor[HTML]{F4F6F8}256                                                      & \multicolumn{1}{r|}{\cellcolor[HTML]{5788B5}1.19}                       & \multicolumn{1}{r|}{\cellcolor[HTML]{5587B4}1.17}                        & \multicolumn{1}{r|}{\cellcolor[HTML]{5D8DB8}1.27}                        & \multicolumn{1}{r|}{\cellcolor[HTML]{83A7C8}1.77}                        & \multicolumn{1}{r|}{\cellcolor[HTML]{FFFFFF}3.56}                        & \multicolumn{1}{r|}{\cellcolor[HTML]{FFFEFE}5.17}                         & \multicolumn{1}{r|}{\cellcolor[HTML]{FFF6F6}18.33}                        & \cellcolor[HTML]{FFFFFF}4.64  \\ \cline{2-10} 
\multirow{-6}{*}{\cellcolor[HTML]{F4F6F8}ROAST}                                        & \cellcolor[HTML]{F4F6F8}512                                                      & \multicolumn{1}{r|}{\cellcolor[HTML]{9DBAD4}2.11}                       & \multicolumn{1}{r|}{\cellcolor[HTML]{8FB0CD}1.92}                        & \multicolumn{1}{r|}{\cellcolor[HTML]{9FBBD4}2.12}                        & \multicolumn{1}{r|}{\cellcolor[HTML]{BED1E2}2.53}                        & \multicolumn{1}{r|}{\cellcolor[HTML]{FFFFFF}4.33}                        & \multicolumn{1}{r|}{\cellcolor[HTML]{FFFEFE}5.98}                         & \multicolumn{1}{r|}{\cellcolor[HTML]{FFF3F3}22.71}                        & \cellcolor[HTML]{FFFEFE}5.96  \\ \hline
\end{tabular}
\caption{Backward pass for different shapes of square weight matrix with input batch of 512. The tile-parameters of multiplication are optimized for each function over "forward +  backward" pass .The measurements are taken with tf32 on A100 (48GB)}
\label{tab:total-bwd}
\end{table}
\begin{table}[h]
\begin{tabular}{|l|l|rrrrrrrrr}
\hline
                                                        &                                                     & \multicolumn{9}{c|}{\textbf{\begin{tabular}[c]{@{}c@{}}update weights (optim.step())(ms) \\ (optimized for forward +  backward)\end{tabular}}}                                                                                                                                                                                                                                                                                                                                                                                                                                                                                                                                       \\ \hline
\rowcolor[HTML]{DFE4EC} 
\multicolumn{1}{|c|}{\cellcolor[HTML]{DFE4EC}\textit{}} & \multicolumn{1}{c|}{\cellcolor[HTML]{DFE4EC}}       & \multicolumn{9}{c|}{\cellcolor[HTML]{DFE4EC}dim (Matrix dimension = dim x dim)}                                                                                                                                                                                                                                                                                                                                                                                                                                                                                                                                                                                                      \\ \hline
\cellcolor[HTML]{DFE4EC}\textit{optim}                  & \cellcolor[HTML]{DFE4EC}\textit{Model}              & \multicolumn{1}{l|}{\cellcolor[HTML]{DFE4EC}{\color[HTML]{333333} \textit{msize}}} & \multicolumn{1}{r|}{\cellcolor[HTML]{8093B3}{\color[HTML]{FFFFFF} 512}} & \multicolumn{1}{r|}{\cellcolor[HTML]{8093B3}{\color[HTML]{FFFFFF} 1024}} & \multicolumn{1}{r|}{\cellcolor[HTML]{8093B3}{\color[HTML]{FFFFFF} 2048}} & \multicolumn{1}{r|}{\cellcolor[HTML]{8093B3}{\color[HTML]{FFFFFF} 4096}} & \multicolumn{1}{r|}{\cellcolor[HTML]{8093B3}{\color[HTML]{FFFFFF} 8096}} & \multicolumn{1}{r|}{\cellcolor[HTML]{8093B3}{\color[HTML]{FFFFFF} 10240}} & \multicolumn{1}{r|}{\cellcolor[HTML]{8093B3}{\color[HTML]{FFFFFF} 20480}} & \multicolumn{1}{l}{Average}                       \\ \cline{1-10}
\cellcolor[HTML]{F4F6F8}adagrad                         & \cellcolor[HTML]{F4F6F8}Full                        & \multicolumn{1}{l|}{\cellcolor[HTML]{F4F6F8}}                                      & \multicolumn{1}{r|}{\cellcolor[HTML]{2968A1}0.14}                       & \multicolumn{1}{r|}{\cellcolor[HTML]{1C5F9B}0.11}                        & \multicolumn{1}{r|}{\cellcolor[HTML]{2D6BA3}0.15}                        & \multicolumn{1}{r|}{\cellcolor[HTML]{F3F6F9}0.60}                        & \multicolumn{1}{r|}{\cellcolor[HTML]{FFEFEF}2.16}                        & \multicolumn{1}{r|}{\cellcolor[HTML]{FFE2E2}3.41}                         & \multicolumn{1}{r|}{\cellcolor[HTML]{FF7676}13.45}                        & \cellcolor[HTML]{FFE8E8}2.86                      \\ \cline{1-10}
\cellcolor[HTML]{F4F6F8}                                & \cellcolor[HTML]{F4F6F8}                            & \multicolumn{1}{r|}{\cellcolor[HTML]{F4F6F8}4}                                     & \multicolumn{1}{r|}{\cellcolor[HTML]{2968A1}0.14}                       & \multicolumn{1}{r|}{\cellcolor[HTML]{1D609C}0.11}                        & \multicolumn{1}{r|}{\cellcolor[HTML]{1D5F9B}0.11}                        & \multicolumn{1}{r|}{\cellcolor[HTML]{1E609C}0.11}                        & \multicolumn{1}{r|}{\cellcolor[HTML]{1E609C}0.11}                        & \multicolumn{1}{r|}{\cellcolor[HTML]{1F619C}0.12}                         & \multicolumn{1}{r|}{\cellcolor[HTML]{D9E4EE}0.54}                         & \cellcolor[HTML]{3A74A8}0.18                      \\ \cline{3-10}
\cellcolor[HTML]{F4F6F8}                                & \cellcolor[HTML]{F4F6F8}                            & \multicolumn{1}{r|}{\cellcolor[HTML]{F4F6F8}32}                                    & \multicolumn{1}{r|}{\cellcolor[HTML]{86AACA}0.35}                       & \multicolumn{1}{r|}{\cellcolor[HTML]{7EA4C6}0.33}                        & \multicolumn{1}{r|}{\cellcolor[HTML]{7DA3C6}0.33}                        & \multicolumn{1}{r|}{\cellcolor[HTML]{7EA4C6}0.33}                        & \multicolumn{1}{r|}{\cellcolor[HTML]{7EA4C6}0.33}                        & \multicolumn{1}{r|}{\cellcolor[HTML]{7FA5C7}0.34}                         & \multicolumn{1}{r|}{\cellcolor[HTML]{87AACA}0.36}                         & \cellcolor[HTML]{81A6C7}0.34                      \\ \cline{3-10}
\cellcolor[HTML]{F4F6F8}                                & \cellcolor[HTML]{F4F6F8}                            & \multicolumn{1}{r|}{\cellcolor[HTML]{F4F6F8}64}                                    & \multicolumn{1}{r|}{\cellcolor[HTML]{F5F8FB}0.61}                       & \multicolumn{1}{r|}{\cellcolor[HTML]{F3F7FA}0.61}                        & \multicolumn{1}{r|}{\cellcolor[HTML]{F4F7FA}0.61}                        & \multicolumn{1}{r|}{\cellcolor[HTML]{F4F7FA}0.61}                        & \multicolumn{1}{r|}{\cellcolor[HTML]{F7F9FB}0.61}                        & \multicolumn{1}{r|}{\cellcolor[HTML]{FAFCFD}0.62}                         & \multicolumn{1}{r|}{\cellcolor[HTML]{F4F7FA}0.61}                         & \cellcolor[HTML]{F5F8FA}0.61                      \\ \cline{3-10}
\rowcolor[HTML]{FFFAFA} 
\cellcolor[HTML]{F4F6F8}                                & \cellcolor[HTML]{F4F6F8}                            & \multicolumn{1}{r|}{\cellcolor[HTML]{F4F6F8}128}                                   & \multicolumn{1}{r|}{\cellcolor[HTML]{FFFAFA}1.15}                       & \multicolumn{1}{r|}{\cellcolor[HTML]{FFFAFA}1.14}                        & \multicolumn{1}{r|}{\cellcolor[HTML]{FFFAFA}1.14}                        & \multicolumn{1}{r|}{\cellcolor[HTML]{FFFAFA}1.14}                        & \multicolumn{1}{r|}{\cellcolor[HTML]{FFFAFA}1.15}                        & \multicolumn{1}{r|}{\cellcolor[HTML]{FFFAFA}1.19}                         & \multicolumn{1}{r|}{\cellcolor[HTML]{FFFAFA}1.18}                         & 1.15                                              \\ \cline{3-10}
\cellcolor[HTML]{F4F6F8}                                & \cellcolor[HTML]{F4F6F8}                            & \multicolumn{1}{r|}{\cellcolor[HTML]{F4F6F8}256}                                   & \multicolumn{1}{r|}{\cellcolor[HTML]{FFEFEF}2.22}                       & \multicolumn{1}{r|}{\cellcolor[HTML]{FFEFEF}2.21}                        & \multicolumn{1}{r|}{\cellcolor[HTML]{FFEFEF}2.21}                        & \multicolumn{1}{r|}{\cellcolor[HTML]{FFEFEF}2.21}                        & \multicolumn{1}{r|}{\cellcolor[HTML]{FFEFEF}2.22}                        & \multicolumn{1}{r|}{\cellcolor[HTML]{FFEEEE}2.26}                         & \multicolumn{1}{r|}{\cellcolor[HTML]{FFDDDD}3.87}                         & \cellcolor[HTML]{FFECEC}2.46                      \\ \cline{3-10}
\multirow{-6}{*}{\cellcolor[HTML]{F4F6F8}}              & \multirow{-6}{*}{\cellcolor[HTML]{F4F6F8}HashedNet} & \multicolumn{1}{r|}{\cellcolor[HTML]{F4F6F8}512}                                   & \multicolumn{1}{r|}{\cellcolor[HTML]{FFD8D8}4.36}                       & \multicolumn{1}{r|}{\cellcolor[HTML]{FFD8D8}4.36}                        & \multicolumn{1}{r|}{\cellcolor[HTML]{FFD8D8}4.35}                        & \multicolumn{1}{r|}{\cellcolor[HTML]{FFD8D8}4.35}                        & \multicolumn{1}{r|}{\cellcolor[HTML]{FFD7D7}4.37}                        & \multicolumn{1}{r|}{\cellcolor[HTML]{FFD7D7}4.40}                         & \multicolumn{1}{r|}{\cellcolor[HTML]{FFD6D6}4.47}                         & \cellcolor[HTML]{FFD7D7}4.38                      \\ \cline{1-10}
\cellcolor[HTML]{F4F6F8}                                & \cellcolor[HTML]{F4F6F8}                            & \multicolumn{1}{r|}{\cellcolor[HTML]{F4F6F8}4}                                     & \multicolumn{1}{r|}{\cellcolor[HTML]{1C5F9B}0.11}                       & \multicolumn{1}{r|}{\cellcolor[HTML]{1C5F9B}0.11}                        & \multicolumn{1}{r|}{\cellcolor[HTML]{1C5F9B}0.11}                        & \multicolumn{1}{r|}{\cellcolor[HTML]{1D609C}0.11}                        & \multicolumn{1}{r|}{\cellcolor[HTML]{20629D}0.12}                        & \multicolumn{1}{r|}{\cellcolor[HTML]{1E609C}0.11}                         & \multicolumn{1}{r|}{\cellcolor[HTML]{1E609C}0.11}                         & \cellcolor[HTML]{1D609C}0.11                      \\ \cline{3-10}
\cellcolor[HTML]{F4F6F8}                                & \cellcolor[HTML]{F4F6F8}                            & \multicolumn{1}{r|}{\cellcolor[HTML]{F4F6F8}32}                                    & \multicolumn{1}{r|}{\cellcolor[HTML]{7EA4C6}0.33}                       & \multicolumn{1}{r|}{\cellcolor[HTML]{7FA5C7}0.34}                        & \multicolumn{1}{r|}{\cellcolor[HTML]{7FA5C7}0.34}                        & \multicolumn{1}{r|}{\cellcolor[HTML]{7CA2C5}0.33}                        & \multicolumn{1}{r|}{\cellcolor[HTML]{7DA3C6}0.33}                        & \multicolumn{1}{r|}{\cellcolor[HTML]{7DA3C6}0.33}                         & \multicolumn{1}{r|}{\cellcolor[HTML]{7EA4C6}0.33}                         & \cellcolor[HTML]{7EA4C6}0.33                      \\ \cline{3-10}
\cellcolor[HTML]{F4F6F8}                                & \cellcolor[HTML]{F4F6F8}                            & \multicolumn{1}{r|}{\cellcolor[HTML]{F4F6F8}64}                                    & \multicolumn{1}{r|}{\cellcolor[HTML]{F3F6F9}0.60}                       & \multicolumn{1}{r|}{\cellcolor[HTML]{F4F7FA}0.61}                        & \multicolumn{1}{r|}{\cellcolor[HTML]{F3F7FA}0.61}                        & \multicolumn{1}{r|}{\cellcolor[HTML]{F3F6F9}0.60}                        & \multicolumn{1}{r|}{\cellcolor[HTML]{F3F7FA}0.61}                        & \multicolumn{1}{r|}{\cellcolor[HTML]{F3F7FA}0.61}                         & \multicolumn{1}{r|}{\cellcolor[HTML]{F4F7FA}0.61}                         & \cellcolor[HTML]{F3F7FA}0.61                      \\ \cline{3-10}
\rowcolor[HTML]{FFFAFA} 
\cellcolor[HTML]{F4F6F8}                                & \cellcolor[HTML]{F4F6F8}                            & \multicolumn{1}{r|}{\cellcolor[HTML]{F4F6F8}128}                                   & \multicolumn{1}{r|}{\cellcolor[HTML]{FFFAFA}1.14}                       & \multicolumn{1}{r|}{\cellcolor[HTML]{FFFAFA}1.14}                        & \multicolumn{1}{r|}{\cellcolor[HTML]{FFFAFA}1.14}                        & \multicolumn{1}{r|}{\cellcolor[HTML]{FFFAFA}1.14}                        & \multicolumn{1}{r|}{\cellcolor[HTML]{FFFAFA}1.14}                        & \multicolumn{1}{r|}{\cellcolor[HTML]{FFFAFA}1.14}                         & \multicolumn{1}{r|}{\cellcolor[HTML]{FFFAFA}1.14}                         & 1.14                                              \\ \cline{3-10}
\rowcolor[HTML]{FFEFEF} 
\cellcolor[HTML]{F4F6F8}                                & \cellcolor[HTML]{F4F6F8}                            & \multicolumn{1}{r|}{\cellcolor[HTML]{F4F6F8}256}                                   & \multicolumn{1}{r|}{\cellcolor[HTML]{FFEFEF}2.21}                       & \multicolumn{1}{r|}{\cellcolor[HTML]{FFEFEF}2.21}                        & \multicolumn{1}{r|}{\cellcolor[HTML]{FFEFEF}2.21}                        & \multicolumn{1}{r|}{\cellcolor[HTML]{FFEFEF}2.21}                        & \multicolumn{1}{r|}{\cellcolor[HTML]{FFEFEF}2.21}                        & \multicolumn{1}{r|}{\cellcolor[HTML]{FFEFEF}2.21}                         & \multicolumn{1}{r|}{\cellcolor[HTML]{FFEFEF}2.21}                         & 2.21                                              \\ \cline{3-10}
\rowcolor[HTML]{FFD8D8} 
\multirow{-6}{*}{\cellcolor[HTML]{F4F6F8}}              & \multirow{-6}{*}{\cellcolor[HTML]{F4F6F8}ROAST}     & \multicolumn{1}{r|}{\cellcolor[HTML]{F4F6F8}512}                                   & \multicolumn{1}{r|}{\cellcolor[HTML]{FFD7D7}4.38}                       & \multicolumn{1}{r|}{\cellcolor[HTML]{FFD8D8}4.35}                        & \multicolumn{1}{r|}{\cellcolor[HTML]{FFD8D8}4.36}                        & \multicolumn{1}{r|}{\cellcolor[HTML]{FFD8D8}4.35}                        & \multicolumn{1}{r|}{\cellcolor[HTML]{FFD8D8}4.35}                        & \multicolumn{1}{r|}{\cellcolor[HTML]{FFD8D8}4.36}                         & \multicolumn{1}{r|}{\cellcolor[HTML]{FFD8D8}4.35}                         & 4.36                                              \\ \hline
\cellcolor[HTML]{F4F6F8}                                & \cellcolor[HTML]{F4F6F8}Full                        & \multicolumn{1}{l|}{\cellcolor[HTML]{F4F6F8}}                                      & \multicolumn{1}{r|}{\cellcolor[HTML]{2F6CA3}0.15}                       & \multicolumn{1}{r|}{\cellcolor[HTML]{2E6CA3}0.15}                        & \multicolumn{1}{r|}{\cellcolor[HTML]{5184B2}0.23}                        & \multicolumn{1}{r|}{\cellcolor[HTML]{FFFBFB}1.06}                        & \multicolumn{1}{r|}{\cellcolor[HTML]{FFDDDD}3.89}                        & \multicolumn{1}{r|}{\cellcolor[HTML]{FFC4C4}6.18}                         & \multicolumn{1}{r|}{\cellcolor[HTML]{FF0000}24.47}                        & \multicolumn{1}{r|}{\cellcolor[HTML]{FFCFCF}5.16} \\ \cline{2-11} 
\cellcolor[HTML]{F4F6F8}                                & \cellcolor[HTML]{F4F6F8}                            & \multicolumn{1}{r|}{\cellcolor[HTML]{F4F6F8}4}                                     & \multicolumn{1}{r|}{\cellcolor[HTML]{306DA4}0.15}                       & \multicolumn{1}{r|}{\cellcolor[HTML]{5285B3}0.23}                        & \multicolumn{1}{r|}{\cellcolor[HTML]{316DA4}0.16}                        & \multicolumn{1}{r|}{\cellcolor[HTML]{326EA5}0.16}                        & \multicolumn{1}{r|}{\cellcolor[HTML]{326EA5}0.16}                        & \multicolumn{1}{r|}{\cellcolor[HTML]{306DA4}0.16}                         & \multicolumn{1}{r|}{\cellcolor[HTML]{3470A6}0.16}                         & \multicolumn{1}{r|}{\cellcolor[HTML]{3671A7}0.17} \\ \cline{3-11} 
\cellcolor[HTML]{F4F6F8}                                & \cellcolor[HTML]{F4F6F8}                            & \multicolumn{1}{r|}{\cellcolor[HTML]{F4F6F8}32}                                    & \multicolumn{1}{r|}{\cellcolor[HTML]{E3EBF2}0.57}                       & \multicolumn{1}{r|}{\cellcolor[HTML]{E6EDF4}0.57}                        & \multicolumn{1}{r|}{\cellcolor[HTML]{E4ECF3}0.57}                        & \multicolumn{1}{r|}{\cellcolor[HTML]{E4ECF3}0.57}                        & \multicolumn{1}{r|}{\cellcolor[HTML]{E3EBF3}0.57}                        & \multicolumn{1}{r|}{\cellcolor[HTML]{E4ECF3}0.57}                         & \multicolumn{1}{r|}{\cellcolor[HTML]{ECF1F6}0.59}                         & \multicolumn{1}{r|}{\cellcolor[HTML]{E5EDF3}0.57} \\ \cline{3-11} 
\rowcolor[HTML]{FFFBFB} 
\cellcolor[HTML]{F4F6F8}                                & \cellcolor[HTML]{F4F6F8}                            & \multicolumn{1}{r|}{\cellcolor[HTML]{F4F6F8}64}                                    & \multicolumn{1}{r|}{\cellcolor[HTML]{FFFBFB}1.06}                       & \multicolumn{1}{r|}{\cellcolor[HTML]{FFFBFB}1.06}                        & \multicolumn{1}{r|}{\cellcolor[HTML]{FFFBFB}1.06}                        & \multicolumn{1}{r|}{\cellcolor[HTML]{FFFBFB}1.06}                        & \multicolumn{1}{r|}{\cellcolor[HTML]{FFFBFB}1.06}                        & \multicolumn{1}{r|}{\cellcolor[HTML]{FFFBFB}1.06}                         & \multicolumn{1}{r|}{\cellcolor[HTML]{FFFAFA}1.16}                         & \multicolumn{1}{r|}{\cellcolor[HTML]{FFFBFB}1.08} \\ \cline{3-11} 
\rowcolor[HTML]{FFF0F0} 
\cellcolor[HTML]{F4F6F8}                                & \cellcolor[HTML]{F4F6F8}                            & \multicolumn{1}{r|}{\cellcolor[HTML]{F4F6F8}128}                                   & \multicolumn{1}{r|}{\cellcolor[HTML]{FFF0F0}2.03}                       & \multicolumn{1}{r|}{\cellcolor[HTML]{FFF0F0}2.05}                        & \multicolumn{1}{r|}{\cellcolor[HTML]{FFF0F0}2.04}                        & \multicolumn{1}{r|}{\cellcolor[HTML]{FFF0F0}2.04}                        & \multicolumn{1}{r|}{\cellcolor[HTML]{FFF0F0}2.05}                        & \multicolumn{1}{r|}{\cellcolor[HTML]{FFF0F0}2.04}                         & \multicolumn{1}{r|}{\cellcolor[HTML]{FFEEEE}2.23}                         & \multicolumn{1}{r|}{\cellcolor[HTML]{FFF0F0}2.07} \\ \cline{3-11} 
\cellcolor[HTML]{F4F6F8}                                & \cellcolor[HTML]{F4F6F8}                            & \multicolumn{1}{r|}{\cellcolor[HTML]{F4F6F8}256}                                   & \multicolumn{1}{r|}{\cellcolor[HTML]{FFDCDC}3.98}                       & \multicolumn{1}{r|}{\cellcolor[HTML]{FFDCDC}3.99}                        & \multicolumn{1}{r|}{\cellcolor[HTML]{FFDCDC}3.98}                        & \multicolumn{1}{r|}{\cellcolor[HTML]{FFDCDC}3.99}                        & \multicolumn{1}{r|}{\cellcolor[HTML]{FFDBDB}4.00}                        & \multicolumn{1}{r|}{\cellcolor[HTML]{FFDBDB}4.00}                         & \multicolumn{1}{r|}{\cellcolor[HTML]{FFD9D9}4.22}                         & \multicolumn{1}{r|}{\cellcolor[HTML]{FFDBDB}4.02} \\ \cline{3-11} 
\rowcolor[HTML]{FFB2B2} 
\cellcolor[HTML]{F4F6F8}                                & \multirow{-6}{*}{\cellcolor[HTML]{F4F6F8}HashedNet} & \multicolumn{1}{r|}{\cellcolor[HTML]{F4F6F8}512}                                   & \multicolumn{1}{r|}{\cellcolor[HTML]{FFB2B2}7.89}                       & \multicolumn{1}{r|}{\cellcolor[HTML]{FFB2B2}7.89}                        & \multicolumn{1}{r|}{\cellcolor[HTML]{FFB2B2}7.89}                        & \multicolumn{1}{r|}{\cellcolor[HTML]{FFB2B2}7.89}                        & \multicolumn{1}{r|}{\cellcolor[HTML]{FFB2B2}7.91}                        & \multicolumn{1}{r|}{\cellcolor[HTML]{FFB2B2}7.90}                         & \multicolumn{1}{r|}{\cellcolor[HTML]{FFAFAF}8.13}                         & \multicolumn{1}{r|}{\cellcolor[HTML]{FFB1B1}7.93} \\ \cline{2-11} 
\cellcolor[HTML]{F4F6F8}                                & \cellcolor[HTML]{F4F6F8}                            & \multicolumn{1}{r|}{\cellcolor[HTML]{F4F6F8}4}                                     & \multicolumn{1}{r|}{\cellcolor[HTML]{306DA4}0.15}                       & \multicolumn{1}{r|}{\cellcolor[HTML]{4F83B2}0.23}                        & \multicolumn{1}{r|}{\cellcolor[HTML]{2F6DA4}0.15}                        & \multicolumn{1}{r|}{\cellcolor[HTML]{316DA4}0.16}                        & \multicolumn{1}{r|}{\cellcolor[HTML]{316DA4}0.16}                        & \multicolumn{1}{r|}{\cellcolor[HTML]{2F6DA4}0.15}                         & \multicolumn{1}{r|}{\cellcolor[HTML]{336FA5}0.16}                         & \multicolumn{1}{r|}{\cellcolor[HTML]{3570A6}0.17} \\ \cline{3-11} 
\cellcolor[HTML]{F4F6F8}                                & \cellcolor[HTML]{F4F6F8}                            & \multicolumn{1}{r|}{\cellcolor[HTML]{F4F6F8}32}                                    & \multicolumn{1}{r|}{\cellcolor[HTML]{E3EBF2}0.57}                       & \multicolumn{1}{r|}{\cellcolor[HTML]{E6EDF4}0.57}                        & \multicolumn{1}{r|}{\cellcolor[HTML]{E4ECF3}0.57}                        & \multicolumn{1}{r|}{\cellcolor[HTML]{E4ECF3}0.57}                        & \multicolumn{1}{r|}{\cellcolor[HTML]{E3EBF3}0.57}                        & \multicolumn{1}{r|}{\cellcolor[HTML]{E3EBF3}0.57}                         & \multicolumn{1}{r|}{\cellcolor[HTML]{E5ECF3}0.57}                         & \multicolumn{1}{r|}{\cellcolor[HTML]{E4ECF3}0.57} \\ \cline{3-11} 
\rowcolor[HTML]{FFFBFB} 
\cellcolor[HTML]{F4F6F8}                                & \cellcolor[HTML]{F4F6F8}                            & \multicolumn{1}{r|}{\cellcolor[HTML]{F4F6F8}64}                                    & \multicolumn{1}{r|}{\cellcolor[HTML]{FFFBFB}1.07}                       & \multicolumn{1}{r|}{\cellcolor[HTML]{FFFBFB}1.06}                        & \multicolumn{1}{r|}{\cellcolor[HTML]{FFFBFB}1.06}                        & \multicolumn{1}{r|}{\cellcolor[HTML]{FFFBFB}1.06}                        & \multicolumn{1}{r|}{\cellcolor[HTML]{FFFBFB}1.06}                        & \multicolumn{1}{r|}{\cellcolor[HTML]{FFFBFB}1.07}                         & \multicolumn{1}{r|}{\cellcolor[HTML]{FFFBFB}1.06}                         & \multicolumn{1}{r|}{\cellcolor[HTML]{FFFBFB}1.06} \\ \cline{3-11} 
\rowcolor[HTML]{FFF0F0} 
\cellcolor[HTML]{F4F6F8}                                & \cellcolor[HTML]{F4F6F8}                            & \multicolumn{1}{r|}{\cellcolor[HTML]{F4F6F8}128}                                   & \multicolumn{1}{r|}{\cellcolor[HTML]{FFF0F0}2.05}                       & \multicolumn{1}{r|}{\cellcolor[HTML]{FFF0F0}2.03}                        & \multicolumn{1}{r|}{\cellcolor[HTML]{FFF0F0}2.04}                        & \multicolumn{1}{r|}{\cellcolor[HTML]{FFF0F0}2.04}                        & \multicolumn{1}{r|}{\cellcolor[HTML]{FFF0F0}2.03}                        & \multicolumn{1}{r|}{\cellcolor[HTML]{FFF0F0}2.04}                         & \multicolumn{1}{r|}{\cellcolor[HTML]{FFF0F0}2.04}                         & \multicolumn{1}{r|}{\cellcolor[HTML]{FFF0F0}2.04} \\ \cline{3-11} 
\rowcolor[HTML]{FFDCDC} 
\cellcolor[HTML]{F4F6F8}                                & \cellcolor[HTML]{F4F6F8}                            & \multicolumn{1}{r|}{\cellcolor[HTML]{F4F6F8}256}                                   & \multicolumn{1}{r|}{\cellcolor[HTML]{FFDBDB}4.01}                       & \multicolumn{1}{r|}{\cellcolor[HTML]{FFDCDC}3.98}                        & \multicolumn{1}{r|}{\cellcolor[HTML]{FFDCDC}3.99}                        & \multicolumn{1}{r|}{\cellcolor[HTML]{FFDCDC}3.99}                        & \multicolumn{1}{r|}{\cellcolor[HTML]{FFDCDC}3.99}                        & \multicolumn{1}{r|}{\cellcolor[HTML]{FFDCDC}3.99}                         & \multicolumn{1}{r|}{\cellcolor[HTML]{FFDCDC}3.99}                         & \multicolumn{1}{r|}{\cellcolor[HTML]{FFDCDC}3.99} \\ \cline{3-11} 
\rowcolor[HTML]{FFB2B2} 
\multirow{-13}{*}{\cellcolor[HTML]{F4F6F8}adam}         & \multirow{-6}{*}{\cellcolor[HTML]{F4F6F8}ROAST}     & \multicolumn{1}{r|}{\cellcolor[HTML]{F4F6F8}512}                                   & \multicolumn{1}{r|}{\cellcolor[HTML]{FFB2B2}7.89}                       & \multicolumn{1}{r|}{\cellcolor[HTML]{FFB2B2}7.89}                        & \multicolumn{1}{r|}{\cellcolor[HTML]{FFB2B2}7.89}                        & \multicolumn{1}{r|}{\cellcolor[HTML]{FFB2B2}7.89}                        & \multicolumn{1}{r|}{\cellcolor[HTML]{FFB2B2}7.89}                        & \multicolumn{1}{r|}{\cellcolor[HTML]{FFB2B2}7.89}                         & \multicolumn{1}{r|}{\cellcolor[HTML]{FFB2B2}7.89}                         & \multicolumn{1}{r|}{\cellcolor[HTML]{FFB2B2}7.89} \\ \hline
\cellcolor[HTML]{F4F6F8}                                & \cellcolor[HTML]{F4F6F8}Full                        & \multicolumn{1}{l|}{\cellcolor[HTML]{F4F6F8}}                                      & \multicolumn{1}{r|}{\cellcolor[HTML]{0E5595}0.08}                       & \multicolumn{1}{r|}{\cellcolor[HTML]{0D5495}0.07}                        & \multicolumn{1}{r|}{\cellcolor[HTML]{0E5595}0.08}                        & \multicolumn{1}{r|}{\cellcolor[HTML]{4279AC}0.20}                        & \multicolumn{1}{r|}{\cellcolor[HTML]{FBFCFD}0.62}                        & \multicolumn{1}{r|}{\cellcolor[HTML]{FFFCFC}0.97}                         & \multicolumn{1}{r|}{\cellcolor[HTML]{FFDCDC}3.92}                         & \multicolumn{1}{r|}{\cellcolor[HTML]{FFFDFD}0.85} \\ \cline{2-11} 
\cellcolor[HTML]{F4F6F8}                                & \cellcolor[HTML]{F4F6F8}                            & \multicolumn{1}{r|}{\cellcolor[HTML]{F4F6F8}4}                                     & \multicolumn{1}{r|}{\cellcolor[HTML]{0F5696}0.08}                       & \multicolumn{1}{r|}{\cellcolor[HTML]{0C5494}0.07}                        & \multicolumn{1}{r|}{\cellcolor[HTML]{0E5595}0.08}                        & \multicolumn{1}{r|}{\cellcolor[HTML]{0D5495}0.07}                        & \multicolumn{1}{r|}{\cellcolor[HTML]{0B5394}0.07}                        & \multicolumn{1}{r|}{\cellcolor[HTML]{0E5595}0.08}                         & \multicolumn{1}{r|}{\cellcolor[HTML]{0F5696}0.08}                         & \multicolumn{1}{r|}{\cellcolor[HTML]{0E5595}0.08} \\ \cline{3-11} 
\cellcolor[HTML]{F4F6F8}                                & \cellcolor[HTML]{F4F6F8}                            & \multicolumn{1}{r|}{\cellcolor[HTML]{F4F6F8}32}                                    & \multicolumn{1}{r|}{\cellcolor[HTML]{21629D}0.12}                       & \multicolumn{1}{r|}{\cellcolor[HTML]{20629D}0.12}                        & \multicolumn{1}{r|}{\cellcolor[HTML]{21639E}0.12}                        & \multicolumn{1}{r|}{\cellcolor[HTML]{22639E}0.12}                        & \multicolumn{1}{r|}{\cellcolor[HTML]{22639E}0.12}                        & \multicolumn{1}{r|}{\cellcolor[HTML]{23649E}0.12}                         & \multicolumn{1}{r|}{\cellcolor[HTML]{3872A7}0.17}                         & \multicolumn{1}{r|}{\cellcolor[HTML]{25659F}0.13} \\ \cline{3-11} 
\cellcolor[HTML]{F4F6F8}                                & \cellcolor[HTML]{F4F6F8}                            & \multicolumn{1}{r|}{\cellcolor[HTML]{F4F6F8}64}                                    & \multicolumn{1}{r|}{\cellcolor[HTML]{4179AC}0.19}                       & \multicolumn{1}{r|}{\cellcolor[HTML]{447BAD}0.20}                        & \multicolumn{1}{r|}{\cellcolor[HTML]{427AAC}0.20}                        & \multicolumn{1}{r|}{\cellcolor[HTML]{427AAC}0.20}                        & \multicolumn{1}{r|}{\cellcolor[HTML]{437AAC}0.20}                        & \multicolumn{1}{r|}{\cellcolor[HTML]{4A7FAF}0.21}                         & \multicolumn{1}{r|}{\cellcolor[HTML]{749DC2}0.31}                         & \multicolumn{1}{r|}{\cellcolor[HTML]{4B80B0}0.22} \\ \cline{3-11} 
\cellcolor[HTML]{F4F6F8}                                & \cellcolor[HTML]{F4F6F8}                            & \multicolumn{1}{r|}{\cellcolor[HTML]{F4F6F8}128}                                   & \multicolumn{1}{r|}{\cellcolor[HTML]{85A9C9}0.35}                       & \multicolumn{1}{r|}{\cellcolor[HTML]{82A7C8}0.34}                        & \multicolumn{1}{r|}{\cellcolor[HTML]{81A6C8}0.34}                        & \multicolumn{1}{r|}{\cellcolor[HTML]{83A7C8}0.35}                        & \multicolumn{1}{r|}{\cellcolor[HTML]{85A9C9}0.35}                        & \multicolumn{1}{r|}{\cellcolor[HTML]{8CAECD}0.37}                         & \multicolumn{1}{r|}{\cellcolor[HTML]{BFD2E3}0.48}                         & \multicolumn{1}{r|}{\cellcolor[HTML]{8DAECD}0.37} \\ \cline{3-11} 
\rowcolor[HTML]{FFFFFF} 
\cellcolor[HTML]{F4F6F8}                                & \cellcolor[HTML]{F4F6F8}                            & \multicolumn{1}{r|}{\cellcolor[HTML]{F4F6F8}256}                                   & \multicolumn{1}{r|}{\cellcolor[HTML]{FFFFFF}0.64}                       & \multicolumn{1}{r|}{\cellcolor[HTML]{FFFFFF}0.64}                        & \multicolumn{1}{r|}{\cellcolor[HTML]{FFFFFF}0.64}                        & \multicolumn{1}{r|}{\cellcolor[HTML]{FFFFFF}0.64}                        & \multicolumn{1}{r|}{\cellcolor[HTML]{FFFFFF}0.65}                        & \multicolumn{1}{r|}{\cellcolor[HTML]{FFFFFF}0.67}                         & \multicolumn{1}{r|}{\cellcolor[HTML]{FFFDFD}0.83}                         & \multicolumn{1}{r|}{\cellcolor[HTML]{FFFFFF}0.67} \\ \cline{3-11} 
\rowcolor[HTML]{FFF9F9} 
\cellcolor[HTML]{F4F6F8}                                & \multirow{-6}{*}{\cellcolor[HTML]{F4F6F8}HashedNet} & \multicolumn{1}{r|}{\cellcolor[HTML]{F4F6F8}512}                                   & \multicolumn{1}{r|}{\cellcolor[HTML]{FFF9F9}1.23}                       & \multicolumn{1}{r|}{\cellcolor[HTML]{FFF9F9}1.23}                        & \multicolumn{1}{r|}{\cellcolor[HTML]{FFF9F9}1.23}                        & \multicolumn{1}{r|}{\cellcolor[HTML]{FFF9F9}1.23}                        & \multicolumn{1}{r|}{\cellcolor[HTML]{FFF9F9}1.25}                        & \multicolumn{1}{r|}{\cellcolor[HTML]{FFF9F9}1.24}                         & \multicolumn{1}{r|}{\cellcolor[HTML]{FFF9F9}1.25}                         & \multicolumn{1}{r|}{\cellcolor[HTML]{FFF9F9}1.24} \\ \cline{2-11} 
\cellcolor[HTML]{F4F6F8}                                & \cellcolor[HTML]{F4F6F8}                            & \multicolumn{1}{r|}{\cellcolor[HTML]{F4F6F8}4}                                     & \multicolumn{1}{r|}{\cellcolor[HTML]{0D5494}0.07}                       & \multicolumn{1}{r|}{\cellcolor[HTML]{0D5494}0.07}                        & \multicolumn{1}{r|}{\cellcolor[HTML]{0D5494}0.07}                        & \multicolumn{1}{r|}{\cellcolor[HTML]{0E5595}0.08}                        & \multicolumn{1}{r|}{\cellcolor[HTML]{0B5394}0.07}                        & \multicolumn{1}{r|}{\cellcolor[HTML]{0D5495}0.07}                         & \multicolumn{1}{r|}{\cellcolor[HTML]{4F83B2}0.23}                         & \multicolumn{1}{r|}{\cellcolor[HTML]{165B99}0.10} \\ \cline{3-11} 
\cellcolor[HTML]{F4F6F8}                                & \cellcolor[HTML]{F4F6F8}                            & \multicolumn{1}{r|}{\cellcolor[HTML]{F4F6F8}32}                                    & \multicolumn{1}{r|}{\cellcolor[HTML]{21629D}0.12}                       & \multicolumn{1}{r|}{\cellcolor[HTML]{20629D}0.12}                        & \multicolumn{1}{r|}{\cellcolor[HTML]{24659F}0.13}                        & \multicolumn{1}{r|}{\cellcolor[HTML]{21629D}0.12}                        & \multicolumn{1}{r|}{\cellcolor[HTML]{21629D}0.12}                        & \multicolumn{1}{r|}{\cellcolor[HTML]{21629D}0.12}                         & \multicolumn{1}{r|}{\cellcolor[HTML]{22639E}0.12}                         & \multicolumn{1}{r|}{\cellcolor[HTML]{21639E}0.12} \\ \cline{3-11} 
\cellcolor[HTML]{F4F6F8}                                & \cellcolor[HTML]{F4F6F8}                            & \multicolumn{1}{r|}{\cellcolor[HTML]{F4F6F8}64}                                    & \multicolumn{1}{r|}{\cellcolor[HTML]{4D82B1}0.22}                       & \multicolumn{1}{r|}{\cellcolor[HTML]{4179AB}0.19}                        & \multicolumn{1}{r|}{\cellcolor[HTML]{437BAC}0.20}                        & \multicolumn{1}{r|}{\cellcolor[HTML]{4179AC}0.19}                        & \multicolumn{1}{r|}{\cellcolor[HTML]{4179AB}0.19}                        & \multicolumn{1}{r|}{\cellcolor[HTML]{427AAC}0.20}                         & \multicolumn{1}{r|}{\cellcolor[HTML]{6A96BE}0.29}                         & \multicolumn{1}{r|}{\cellcolor[HTML]{497FAF}0.21} \\ \cline{3-11} 
\cellcolor[HTML]{F4F6F8}                                & \cellcolor[HTML]{F4F6F8}                            & \multicolumn{1}{r|}{\cellcolor[HTML]{F4F6F8}128}                                   & \multicolumn{1}{r|}{\cellcolor[HTML]{82A7C8}0.34}                       & \multicolumn{1}{r|}{\cellcolor[HTML]{83A7C8}0.35}                        & \multicolumn{1}{r|}{\cellcolor[HTML]{82A7C8}0.34}                        & \multicolumn{1}{r|}{\cellcolor[HTML]{82A7C8}0.34}                        & \multicolumn{1}{r|}{\cellcolor[HTML]{81A6C8}0.34}                        & \multicolumn{1}{r|}{\cellcolor[HTML]{85A9C9}0.35}                         & \multicolumn{1}{r|}{\cellcolor[HTML]{99B7D2}0.40}                         & \multicolumn{1}{r|}{\cellcolor[HTML]{86A9CA}0.35} \\ \cline{3-11} 
\rowcolor[HTML]{FFFFFF} 
\cellcolor[HTML]{F4F6F8}                                & \cellcolor[HTML]{F4F6F8}                            & \multicolumn{1}{r|}{\cellcolor[HTML]{F4F6F8}256}                                   & \multicolumn{1}{r|}{\cellcolor[HTML]{FFFFFF}0.64}                       & \multicolumn{1}{r|}{\cellcolor[HTML]{FFFFFF}0.65}                        & \multicolumn{1}{r|}{\cellcolor[HTML]{FFFFFF}0.64}                        & \multicolumn{1}{r|}{\cellcolor[HTML]{FFFFFF}0.64}                        & \multicolumn{1}{r|}{\cellcolor[HTML]{FFFFFF}0.64}                        & \multicolumn{1}{r|}{\cellcolor[HTML]{FFFFFF}0.65}                         & \multicolumn{1}{r|}{\cellcolor[HTML]{FFFFFF}0.64}                         & \multicolumn{1}{r|}{\cellcolor[HTML]{FFFFFF}0.64} \\ \cline{3-11} 
\rowcolor[HTML]{FFF9F9} 
\multirow{-13}{*}{\cellcolor[HTML]{F4F6F8}sgd}          & \multirow{-6}{*}{\cellcolor[HTML]{F4F6F8}ROAST}     & \multicolumn{1}{r|}{\cellcolor[HTML]{F4F6F8}512}                                   & \multicolumn{1}{r|}{\cellcolor[HTML]{FFF9F9}1.27}                       & \multicolumn{1}{r|}{\cellcolor[HTML]{FFF9F9}1.23}                        & \multicolumn{1}{r|}{\cellcolor[HTML]{FFF9F9}1.23}                        & \multicolumn{1}{r|}{\cellcolor[HTML]{FFF9F9}1.23}                        & \multicolumn{1}{r|}{\cellcolor[HTML]{FFF9F9}1.28}                        & \multicolumn{1}{r|}{\cellcolor[HTML]{FFF9F9}1.23}                         & \multicolumn{1}{r|}{\cellcolor[HTML]{FFF5F5}1.62}                         & \multicolumn{1}{r|}{\cellcolor[HTML]{FFF8F8}1.30} \\ \hline
\end{tabular}
\caption{Weight update operation (optimizer.step()) for different shapes of square weight matrix with input batch of 512. The tile-parameters of multiplication are optimized for each function over "forward +  backward" pass .The measurements are taken with tf32 on A100 (48GB)}
\label{tab:total-opt}
\end{table}
\begin{table}[]
\begin{tabular}{|l|l|rrrrrrrrr}
\hline
                                                        &                                                     & \multicolumn{9}{c|}{\textbf{\begin{tabular}[c]{@{}c@{}}total = fwd + bkwd + optimize (ms) \\ (optimized for forward +  backward)\end{tabular}}}                                                                                                                                                                                                                                                                                                                                                                                                                                                                                                                                        \\ \hline
\rowcolor[HTML]{DFE4EC} 
\multicolumn{1}{|c|}{\cellcolor[HTML]{DFE4EC}\textit{}} & \multicolumn{1}{c|}{\cellcolor[HTML]{DFE4EC}}       & \multicolumn{9}{c|}{\cellcolor[HTML]{DFE4EC}dim (Matrix dimension = dim x dim)}                                                                                                                                                                                                                                                                                                                                                                                                                                                                                                                                                                                                        \\ \hline
\cellcolor[HTML]{DFE4EC}\textit{optim}                  & \cellcolor[HTML]{DFE4EC}\textit{Model}              & \multicolumn{1}{l|}{\cellcolor[HTML]{DFE4EC}{\color[HTML]{333333} \textit{msize}}} & \multicolumn{1}{r|}{\cellcolor[HTML]{8093B3}{\color[HTML]{FFFFFF} 512}} & \multicolumn{1}{r|}{\cellcolor[HTML]{8093B3}{\color[HTML]{FFFFFF} 1024}} & \multicolumn{1}{r|}{\cellcolor[HTML]{8093B3}{\color[HTML]{FFFFFF} 2048}} & \multicolumn{1}{r|}{\cellcolor[HTML]{8093B3}{\color[HTML]{FFFFFF} 4096}} & \multicolumn{1}{r|}{\cellcolor[HTML]{8093B3}{\color[HTML]{FFFFFF} 8096}} & \multicolumn{1}{r|}{\cellcolor[HTML]{8093B3}{\color[HTML]{FFFFFF} 10240}} & \multicolumn{1}{r|}{\cellcolor[HTML]{8093B3}{\color[HTML]{FFFFFF} 20480}} & \multicolumn{1}{l}{Average}                         \\ \cline{1-10}
\cellcolor[HTML]{F4F6F8}adagrad                         & \cellcolor[HTML]{F4F6F8}Full                        & \multicolumn{1}{l|}{\cellcolor[HTML]{F4F6F8}}                                      & \multicolumn{1}{r|}{\cellcolor[HTML]{135997}0.65}                       & \multicolumn{1}{r|}{\cellcolor[HTML]{0B5394}0.46}                        & \multicolumn{1}{r|}{\cellcolor[HTML]{0E5595}0.51}                        & \multicolumn{1}{r|}{\cellcolor[HTML]{2F6CA3}1.32}                        & \multicolumn{1}{r|}{\cellcolor[HTML]{A2BDD6}4.17}                        & \multicolumn{1}{r|}{\cellcolor[HTML]{FAFBFC}6.33}                         & \multicolumn{1}{r|}{\cellcolor[HTML]{FFF9F9}24.13}                        & \cellcolor[HTML]{D3DFEB}5.37                        \\ \cline{1-10}
\cellcolor[HTML]{F4F6F8}                                & \cellcolor[HTML]{F4F6F8}                            & \multicolumn{1}{r|}{\cellcolor[HTML]{F4F6F8}4}                                     & \multicolumn{1}{r|}{\cellcolor[HTML]{2867A0}1.16}                       & \multicolumn{1}{r|}{\cellcolor[HTML]{21629D}0.99}                        & \multicolumn{1}{r|}{\cellcolor[HTML]{3E77AA}1.71}                        & \multicolumn{1}{r|}{\cellcolor[HTML]{B9CEE0}4.74}                        & \multicolumn{1}{r|}{\cellcolor[HTML]{FFFCFC}14.86}                       & \multicolumn{1}{r|}{\cellcolor[HTML]{FFF9F9}22.95}                        & \multicolumn{1}{r|}{\cellcolor[HTML]{FFDEDE}92.78}                        & \cellcolor[HTML]{FFFAFA}19.88                       \\ \cline{3-10}
\cellcolor[HTML]{F4F6F8}                                & \cellcolor[HTML]{F4F6F8}                            & \multicolumn{1}{r|}{\cellcolor[HTML]{F4F6F8}32}                                    & \multicolumn{1}{r|}{\cellcolor[HTML]{336FA5}1.43}                       & \multicolumn{1}{r|}{\cellcolor[HTML]{346FA5}1.44}                        & \multicolumn{1}{r|}{\cellcolor[HTML]{749DC2}3.03}                        & \multicolumn{1}{r|}{\cellcolor[HTML]{FFFEFE}10.37}                       & \multicolumn{1}{r|}{\cellcolor[HTML]{FFF3F3}38.19}                       & \multicolumn{1}{r|}{\cellcolor[HTML]{FFEAEA}61.35}                        & \multicolumn{1}{r|}{\cellcolor[HTML]{FFA1A1}253.72}                       & \cellcolor[HTML]{FFEEEE}52.79                       \\ \cline{3-10}
\cellcolor[HTML]{F4F6F8}                                & \cellcolor[HTML]{F4F6F8}                            & \multicolumn{1}{r|}{\cellcolor[HTML]{F4F6F8}64}                                    & \multicolumn{1}{r|}{\cellcolor[HTML]{3D76AA}1.68}                       & \multicolumn{1}{r|}{\cellcolor[HTML]{5083B2}2.14}                        & \multicolumn{1}{r|}{\cellcolor[HTML]{B0C8DC}4.53}                        & \multicolumn{1}{r|}{\cellcolor[HTML]{FFFBFB}17.83}                       & \multicolumn{1}{r|}{\cellcolor[HTML]{FFE8E8}67.74}                       & \multicolumn{1}{r|}{\cellcolor[HTML]{FFD8D8}110.15}                       & \multicolumn{1}{r|}{\cellcolor[HTML]{FF5555}450.66}                       & \cellcolor[HTML]{FFDEDE}93.53                       \\ \cline{3-10}
\cellcolor[HTML]{F4F6F8}                                & \cellcolor[HTML]{F4F6F8}                            & \multicolumn{1}{r|}{\cellcolor[HTML]{F4F6F8}128}                                   & \multicolumn{1}{r|}{\cellcolor[HTML]{5889B5}2.34}                       & \multicolumn{1}{r|}{\cellcolor[HTML]{749DC2}3.04}                        & \multicolumn{1}{r|}{\cellcolor[HTML]{F2F6F9}6.15}                        & \multicolumn{1}{r|}{\cellcolor[HTML]{FFF9F9}22.62}                       & \multicolumn{1}{r|}{\cellcolor[HTML]{FFE1E1}86.36}                       & \multicolumn{1}{r|}{\cellcolor[HTML]{FFCDCD}138.51}                       & \multicolumn{1}{r|}{\cellcolor[HTML]{FF2828}567.83}                       & \cellcolor[HTML]{FFD5D5}118.12                      \\ \cline{3-10}
\cellcolor[HTML]{F4F6F8}                                & \cellcolor[HTML]{F4F6F8}                            & \multicolumn{1}{r|}{\cellcolor[HTML]{F4F6F8}256}                                   & \multicolumn{1}{r|}{\cellcolor[HTML]{8FB0CE}3.71}                       & \multicolumn{1}{r|}{\cellcolor[HTML]{B3C9DD}4.59}                        & \multicolumn{1}{r|}{\cellcolor[HTML]{FFFFFF}7.95}                        & \multicolumn{1}{r|}{\cellcolor[HTML]{FFF8F8}26.73}                       & \multicolumn{1}{r|}{\cellcolor[HTML]{FFDCDC}98.42}                       & \multicolumn{1}{r|}{\cellcolor[HTML]{FFC6C6}155.92}                       & \multicolumn{1}{r|}{\cellcolor[HTML]{FF0D0D}638.80}                       & \cellcolor[HTML]{FFCFCF}133.73                      \\ \cline{3-10}
\multirow{-6}{*}{\cellcolor[HTML]{F4F6F8}}              & \multirow{-6}{*}{\cellcolor[HTML]{F4F6F8}HashedNet} & \multicolumn{1}{r|}{\cellcolor[HTML]{F4F6F8}512}                                   & \multicolumn{1}{r|}{\cellcolor[HTML]{FFFFFF}6.58}                       & \multicolumn{1}{r|}{\cellcolor[HTML]{FFFFFF}7.47}                        & \multicolumn{1}{r|}{\cellcolor[HTML]{FFFEFE}11.13}                       & \multicolumn{1}{r|}{\cellcolor[HTML]{FFF6F6}30.96}                       & \multicolumn{1}{r|}{\cellcolor[HTML]{FFD9D9}107.21}                      & \multicolumn{1}{r|}{\cellcolor[HTML]{FFC2C2}168.30}                       & \multicolumn{1}{r|}{\cellcolor[HTML]{FF0202}668.12}                       & \cellcolor[HTML]{FFCBCB}142.83                      \\ \cline{1-10}
\cellcolor[HTML]{F4F6F8}                                & \cellcolor[HTML]{F4F6F8}                            & \multicolumn{1}{r|}{\cellcolor[HTML]{F4F6F8}4}                                     & \multicolumn{1}{r|}{\cellcolor[HTML]{20619D}0.95}                       & \multicolumn{1}{r|}{\cellcolor[HTML]{20619D}0.95}                        & \multicolumn{1}{r|}{\cellcolor[HTML]{22639E}1.02}                        & \multicolumn{1}{r|}{\cellcolor[HTML]{427AAC}1.81}                        & \multicolumn{1}{r|}{\cellcolor[HTML]{9FBBD5}4.09}                        & \multicolumn{1}{r|}{\cellcolor[HTML]{E6EDF4}5.84}                         & \multicolumn{1}{r|}{\cellcolor[HTML]{FFFAFA}21.64}                        & \cellcolor[HTML]{CBDAE8}5.19                        \\ \cline{3-10}
\cellcolor[HTML]{F4F6F8}                                & \cellcolor[HTML]{F4F6F8}                            & \multicolumn{1}{r|}{\cellcolor[HTML]{F4F6F8}32}                                    & \multicolumn{1}{r|}{\cellcolor[HTML]{2A69A1}1.21}                       & \multicolumn{1}{r|}{\cellcolor[HTML]{2D6AA2}1.27}                        & \multicolumn{1}{r|}{\cellcolor[HTML]{316DA4}1.38}                        & \multicolumn{1}{r|}{\cellcolor[HTML]{487EAE}1.94}                        & \multicolumn{1}{r|}{\cellcolor[HTML]{AFC6DC}4.49}                        & \multicolumn{1}{r|}{\cellcolor[HTML]{FFFFFF}6.46}                         & \multicolumn{1}{r|}{\cellcolor[HTML]{FFF9F9}23.00}                        & \cellcolor[HTML]{DFE8F1}5.68                        \\ \cline{3-10}
\cellcolor[HTML]{F4F6F8}                                & \cellcolor[HTML]{F4F6F8}                            & \multicolumn{1}{r|}{\cellcolor[HTML]{F4F6F8}64}                                    & \multicolumn{1}{r|}{\cellcolor[HTML]{3772A7}1.54}                       & \multicolumn{1}{r|}{\cellcolor[HTML]{3B75A9}1.64}                        & \multicolumn{1}{r|}{\cellcolor[HTML]{3E77AA}1.70}                        & \multicolumn{1}{r|}{\cellcolor[HTML]{5889B5}2.34}                        & \multicolumn{1}{r|}{\cellcolor[HTML]{BED1E2}4.87}                        & \multicolumn{1}{r|}{\cellcolor[HTML]{FFFFFF}6.86}                         & \multicolumn{1}{r|}{\cellcolor[HTML]{FFF9F9}23.90}                        & \cellcolor[HTML]{F1F5F9}6.12                        \\ \cline{3-10}
\cellcolor[HTML]{F4F6F8}                                & \cellcolor[HTML]{F4F6F8}                            & \multicolumn{1}{r|}{\cellcolor[HTML]{F4F6F8}128}                                   & \multicolumn{1}{r|}{\cellcolor[HTML]{5184B3}2.18}                       & \multicolumn{1}{r|}{\cellcolor[HTML]{5386B3}2.22}                        & \multicolumn{1}{r|}{\cellcolor[HTML]{5788B5}2.31}                        & \multicolumn{1}{r|}{\cellcolor[HTML]{759EC2}3.06}                        & \multicolumn{1}{r|}{\cellcolor[HTML]{E1E9F1}5.72}                        & \multicolumn{1}{r|}{\cellcolor[HTML]{FFFFFF}7.97}                         & \multicolumn{1}{r|}{\cellcolor[HTML]{FFF8F8}27.28}                        & \cellcolor[HTML]{FFFFFF}7.25                        \\ \cline{3-10}
\cellcolor[HTML]{F4F6F8}                                & \cellcolor[HTML]{F4F6F8}                            & \multicolumn{1}{r|}{\cellcolor[HTML]{F4F6F8}256}                                   & \multicolumn{1}{r|}{\cellcolor[HTML]{8AACCB}3.57}                       & \multicolumn{1}{r|}{\cellcolor[HTML]{89ACCB}3.56}                        & \multicolumn{1}{r|}{\cellcolor[HTML]{8EAFCD}3.67}                        & \multicolumn{1}{r|}{\cellcolor[HTML]{ADC5DB}4.43}                        & \multicolumn{1}{r|}{\cellcolor[HTML]{FFFFFF}7.10}                        & \multicolumn{1}{r|}{\cellcolor[HTML]{FFFEFE}9.40}                         & \multicolumn{1}{r|}{\cellcolor[HTML]{FFF7F7}28.35}                        & \cellcolor[HTML]{FFFFFF}8.58                        \\ \cline{3-10}
\multirow{-6}{*}{\cellcolor[HTML]{F4F6F8}}              & \multirow{-6}{*}{\cellcolor[HTML]{F4F6F8}ROAST}     & \multicolumn{1}{r|}{\cellcolor[HTML]{F4F6F8}512}                                   & \multicolumn{1}{r|}{\cellcolor[HTML]{FFFFFF}6.70}                       & \multicolumn{1}{r|}{\cellcolor[HTML]{F9FBFC}6.32}                        & \multicolumn{1}{r|}{\cellcolor[HTML]{FFFFFF}6.62}                        & \multicolumn{1}{r|}{\cellcolor[HTML]{FFFFFF}7.29}                        & \multicolumn{1}{r|}{\cellcolor[HTML]{FFFEFE}9.97}                        & \multicolumn{1}{r|}{\cellcolor[HTML]{FFFDFD}12.31}                        & \multicolumn{1}{r|}{\cellcolor[HTML]{FFF6F6}32.04}                        & \cellcolor[HTML]{FFFEFE}11.61                       \\ \hline
\cellcolor[HTML]{F4F6F8}                                & \cellcolor[HTML]{F4F6F8}Full                        & \multicolumn{1}{l|}{\cellcolor[HTML]{F4F6F8}}                                      & \multicolumn{1}{r|}{\cellcolor[HTML]{0D5495}0.50}                       & \multicolumn{1}{r|}{\cellcolor[HTML]{0C5494}0.48}                        & \multicolumn{1}{r|}{\cellcolor[HTML]{115796}0.60}                        & \multicolumn{1}{r|}{\cellcolor[HTML]{4179AB}1.78}                        & \multicolumn{1}{r|}{\cellcolor[HTML]{E8EFF5}5.89}                        & \multicolumn{1}{r|}{\cellcolor[HTML]{FFFEFE}9.11}                         & \multicolumn{1}{r|}{\cellcolor[HTML]{FFF5F5}35.01}                        & \multicolumn{1}{r|}{\cellcolor[HTML]{FFFFFF}7.62}   \\ \cline{2-11} 
\cellcolor[HTML]{F4F6F8}                                & \cellcolor[HTML]{F4F6F8}                            & \multicolumn{1}{r|}{\cellcolor[HTML]{F4F6F8}4}                                     & \multicolumn{1}{r|}{\cellcolor[HTML]{21639E}1.00}                       & \multicolumn{1}{r|}{\cellcolor[HTML]{3873A8}1.56}                        & \multicolumn{1}{r|}{\cellcolor[HTML]{4078AB}1.76}                        & \multicolumn{1}{r|}{\cellcolor[HTML]{BCD0E1}4.81}                        & \multicolumn{1}{r|}{\cellcolor[HTML]{FFFCFC}14.94}                       & \multicolumn{1}{r|}{\cellcolor[HTML]{FFF9F9}23.07}                        & \multicolumn{1}{r|}{\cellcolor[HTML]{FFE1E1}86.76}                        & \multicolumn{1}{r|}{\cellcolor[HTML]{FFFBFB}19.13}  \\ \cline{3-11} 
\cellcolor[HTML]{F4F6F8}                                & \cellcolor[HTML]{F4F6F8}                            & \multicolumn{1}{r|}{\cellcolor[HTML]{F4F6F8}32}                                    & \multicolumn{1}{r|}{\cellcolor[HTML]{336FA5}1.43}                       & \multicolumn{1}{r|}{\cellcolor[HTML]{4179AB}1.78}                        & \multicolumn{1}{r|}{\cellcolor[HTML]{7EA4C6}3.29}                        & \multicolumn{1}{r|}{\cellcolor[HTML]{FFFEFE}10.60}                       & \multicolumn{1}{r|}{\cellcolor[HTML]{FFF3F3}38.45}                       & \multicolumn{1}{r|}{\cellcolor[HTML]{FFEAEA}61.64}                        & \multicolumn{1}{r|}{\cellcolor[HTML]{FFA1A1}253.20}                       & \multicolumn{1}{r|}{\cellcolor[HTML]{FFEEEE}52.91}  \\ \cline{3-11} 
\cellcolor[HTML]{F4F6F8}                                & \cellcolor[HTML]{F4F6F8}                            & \multicolumn{1}{r|}{\cellcolor[HTML]{F4F6F8}64}                                    & \multicolumn{1}{r|}{\cellcolor[HTML]{4B80B0}2.03}                       & \multicolumn{1}{r|}{\cellcolor[HTML]{6391BA}2.63}                        & \multicolumn{1}{r|}{\cellcolor[HTML]{C2D4E4}4.97}                        & \multicolumn{1}{r|}{\cellcolor[HTML]{FFFBFB}18.35}                       & \multicolumn{1}{r|}{\cellcolor[HTML]{FFE8E8}68.28}                       & \multicolumn{1}{r|}{\cellcolor[HTML]{FFD8D8}110.63}                       & \multicolumn{1}{r|}{\cellcolor[HTML]{FF5555}450.86}                       & \multicolumn{1}{r|}{\cellcolor[HTML]{FFDEDE}93.96}  \\ \cline{3-11} 
\cellcolor[HTML]{F4F6F8}                                & \cellcolor[HTML]{F4F6F8}                            & \multicolumn{1}{r|}{\cellcolor[HTML]{F4F6F8}128}                                   & \multicolumn{1}{r|}{\cellcolor[HTML]{7AA1C4}3.18}                       & \multicolumn{1}{r|}{\cellcolor[HTML]{A6C0D8}4.27}                        & \multicolumn{1}{r|}{\cellcolor[HTML]{FFFFFF}7.02}                        & \multicolumn{1}{r|}{\cellcolor[HTML]{FFF9F9}23.54}                       & \multicolumn{1}{r|}{\cellcolor[HTML]{FFE0E0}87.47}                       & \multicolumn{1}{r|}{\cellcolor[HTML]{FFCDCD}139.30}                       & \multicolumn{1}{r|}{\cellcolor[HTML]{FF2828}568.72}                       & \multicolumn{1}{r|}{\cellcolor[HTML]{FFD4D4}119.07} \\ \cline{3-11} 
\cellcolor[HTML]{F4F6F8}                                & \cellcolor[HTML]{F4F6F8}                            & \multicolumn{1}{r|}{\cellcolor[HTML]{F4F6F8}256}                                   & \multicolumn{1}{r|}{\cellcolor[HTML]{D6E2ED}5.45}                       & \multicolumn{1}{r|}{\cellcolor[HTML]{F8FAFC}6.30}                        & \multicolumn{1}{r|}{\cellcolor[HTML]{FFFEFE}9.71}                        & \multicolumn{1}{r|}{\cellcolor[HTML]{FFF7F7}28.66}                       & \multicolumn{1}{r|}{\cellcolor[HTML]{FFDCDC}100.19}                      & \multicolumn{1}{r|}{\cellcolor[HTML]{FFC6C6}157.55}                       & \multicolumn{1}{r|}{\cellcolor[HTML]{FF0F0F}633.80}                       & \multicolumn{1}{r|}{\cellcolor[HTML]{FFCECE}134.52} \\ \cline{3-11} 
\cellcolor[HTML]{F4F6F8}                                & \multirow{-6}{*}{\cellcolor[HTML]{F4F6F8}HashedNet} & \multicolumn{1}{r|}{\cellcolor[HTML]{F4F6F8}512}                                   & \multicolumn{1}{r|}{\cellcolor[HTML]{FFFEFE}10.08}                      & \multicolumn{1}{r|}{\cellcolor[HTML]{FFFEFE}10.94}                       & \multicolumn{1}{r|}{\cellcolor[HTML]{FFFCFC}14.64}                       & \multicolumn{1}{r|}{\cellcolor[HTML]{FFF5F5}34.56}                       & \multicolumn{1}{r|}{\cellcolor[HTML]{FFD8D8}110.71}                      & \multicolumn{1}{r|}{\cellcolor[HTML]{FFC0C0}171.67}                       & \multicolumn{1}{r|}{\cellcolor[HTML]{FF0000}672.24}                       & \multicolumn{1}{r|}{\cellcolor[HTML]{FFCACA}146.41} \\ \cline{2-11} 
\cellcolor[HTML]{F4F6F8}                                & \cellcolor[HTML]{F4F6F8}                            & \multicolumn{1}{r|}{\cellcolor[HTML]{F4F6F8}4}                                     & \multicolumn{1}{r|}{\cellcolor[HTML]{21639E}1.00}                       & \multicolumn{1}{r|}{\cellcolor[HTML]{2C6AA2}1.27}                        & \multicolumn{1}{r|}{\cellcolor[HTML]{24649E}1.05}                        & \multicolumn{1}{r|}{\cellcolor[HTML]{447BAD}1.86}                        & \multicolumn{1}{r|}{\cellcolor[HTML]{9EBAD4}4.06}                        & \multicolumn{1}{r|}{\cellcolor[HTML]{E8EEF4}5.89}                         & \multicolumn{1}{r|}{\cellcolor[HTML]{FFFAFA}21.71}                        & \multicolumn{1}{r|}{\cellcolor[HTML]{CEDCE9}5.26}   \\ \cline{3-11} 
\cellcolor[HTML]{F4F6F8}                                & \cellcolor[HTML]{F4F6F8}                            & \multicolumn{1}{r|}{\cellcolor[HTML]{F4F6F8}32}                                    & \multicolumn{1}{r|}{\cellcolor[HTML]{3470A6}1.45}                       & \multicolumn{1}{r|}{\cellcolor[HTML]{3873A8}1.56}                        & \multicolumn{1}{r|}{\cellcolor[HTML]{3772A7}1.52}                        & \multicolumn{1}{r|}{\cellcolor[HTML]{5385B3}2.21}                        & \multicolumn{1}{r|}{\cellcolor[HTML]{B8CDE0}4.72}                        & \multicolumn{1}{r|}{\cellcolor[HTML]{FFFFFF}6.69}                         & \multicolumn{1}{r|}{\cellcolor[HTML]{FFF9F9}23.28}                        & \multicolumn{1}{r|}{\cellcolor[HTML]{E9EFF5}5.92}   \\ \cline{3-11} 
\cellcolor[HTML]{F4F6F8}                                & \cellcolor[HTML]{F4F6F8}                            & \multicolumn{1}{r|}{\cellcolor[HTML]{F4F6F8}64}                                    & \multicolumn{1}{r|}{\cellcolor[HTML]{4F83B2}2.13}                       & \multicolumn{1}{r|}{\cellcolor[HTML]{4B80B0}2.02}                        & \multicolumn{1}{r|}{\cellcolor[HTML]{5184B3}2.18}                        & \multicolumn{1}{r|}{\cellcolor[HTML]{6A96BE}2.80}                        & \multicolumn{1}{r|}{\cellcolor[HTML]{D1DFEB}5.34}                        & \multicolumn{1}{r|}{\cellcolor[HTML]{FFFFFF}7.39}                         & \multicolumn{1}{r|}{\cellcolor[HTML]{FFF9F9}24.35}                        & \multicolumn{1}{r|}{\cellcolor[HTML]{FFFFFF}6.60}   \\ \cline{3-11} 
\cellcolor[HTML]{F4F6F8}                                & \cellcolor[HTML]{F4F6F8}                            & \multicolumn{1}{r|}{\cellcolor[HTML]{F4F6F8}128}                                   & \multicolumn{1}{r|}{\cellcolor[HTML]{7DA3C6}3.26}                       & \multicolumn{1}{r|}{\cellcolor[HTML]{779FC3}3.11}                        & \multicolumn{1}{r|}{\cellcolor[HTML]{7CA2C5}3.23}                        & \multicolumn{1}{r|}{\cellcolor[HTML]{99B7D2}3.95}                        & \multicolumn{1}{r|}{\cellcolor[HTML]{FFFFFF}6.62}                        & \multicolumn{1}{r|}{\cellcolor[HTML]{FFFFFF}8.85}                         & \multicolumn{1}{r|}{\cellcolor[HTML]{FFF7F7}28.22}                        & \multicolumn{1}{r|}{\cellcolor[HTML]{FFFFFF}8.18}   \\ \cline{3-11} 
\cellcolor[HTML]{F4F6F8}                                & \cellcolor[HTML]{F4F6F8}                            & \multicolumn{1}{r|}{\cellcolor[HTML]{F4F6F8}256}                                   & \multicolumn{1}{r|}{\cellcolor[HTML]{E5ECF3}5.82}                       & \multicolumn{1}{r|}{\cellcolor[HTML]{D1DEEB}5.33}                        & \multicolumn{1}{r|}{\cellcolor[HTML]{D6E2ED}5.45}                        & \multicolumn{1}{r|}{\cellcolor[HTML]{F5F8FA}6.21}                        & \multicolumn{1}{r|}{\cellcolor[HTML]{FFFFFF}8.97}                        & \multicolumn{1}{r|}{\cellcolor[HTML]{FFFEFE}11.15}                        & \multicolumn{1}{r|}{\cellcolor[HTML]{FFF6F6}30.19}                        & \multicolumn{1}{r|}{\cellcolor[HTML]{FFFEFE}10.45}  \\ \cline{3-11} 
\multirow{-13}{*}{\cellcolor[HTML]{F4F6F8}adam}         & \multirow{-6}{*}{\cellcolor[HTML]{F4F6F8}ROAST}     & \multicolumn{1}{r|}{\cellcolor[HTML]{F4F6F8}512}                                   & \multicolumn{1}{r|}{\cellcolor[HTML]{FFFEFE}9.82}                       & \multicolumn{1}{r|}{\cellcolor[HTML]{FFFEFE}9.87}                        & \multicolumn{1}{r|}{\cellcolor[HTML]{FFFEFE}10.14}                       & \multicolumn{1}{r|}{\cellcolor[HTML]{FFFEFE}10.90}                       & \multicolumn{1}{r|}{\cellcolor[HTML]{FFFDFD}13.52}                       & \multicolumn{1}{r|}{\cellcolor[HTML]{FFFCFC}15.82}                        & \multicolumn{1}{r|}{\cellcolor[HTML]{FFF4F4}35.59}                        & \multicolumn{1}{r|}{\cellcolor[HTML]{FFFCFC}15.09}  \\ \hline
\cellcolor[HTML]{F4F6F8}                                & \cellcolor[HTML]{F4F6F8}Full                        & \multicolumn{1}{l|}{\cellcolor[HTML]{F4F6F8}}                                      & \multicolumn{1}{r|}{\cellcolor[HTML]{0B5394}0.44}                       & \multicolumn{1}{r|}{\cellcolor[HTML]{0B5394}0.43}                        & \multicolumn{1}{r|}{\cellcolor[HTML]{0C5394}0.46}                        & \multicolumn{1}{r|}{\cellcolor[HTML]{1E609C}0.90}                        & \multicolumn{1}{r|}{\cellcolor[HTML]{6391BA}2.62}                        & \multicolumn{1}{r|}{\cellcolor[HTML]{97B6D1}3.90}                         & \multicolumn{1}{r|}{\cellcolor[HTML]{FFFCFC}14.68}                        & \multicolumn{1}{r|}{\cellcolor[HTML]{81A6C7}3.35}   \\ \cline{2-11} 
\cellcolor[HTML]{F4F6F8}                                & \cellcolor[HTML]{F4F6F8}                            & \multicolumn{1}{r|}{\cellcolor[HTML]{F4F6F8}4}                                     & \multicolumn{1}{r|}{\cellcolor[HTML]{2C6AA2}1.25}                       & \multicolumn{1}{r|}{\cellcolor[HTML]{20619D}0.95}                        & \multicolumn{1}{r|}{\cellcolor[HTML]{3E77AA}1.70}                        & \multicolumn{1}{r|}{\cellcolor[HTML]{B8CDE0}4.72}                        & \multicolumn{1}{r|}{\cellcolor[HTML]{FFFCFC}14.76}                       & \multicolumn{1}{r|}{\cellcolor[HTML]{FFF9F9}22.96}                        & \multicolumn{1}{r|}{\cellcolor[HTML]{FFE1E1}86.70}                        & \multicolumn{1}{r|}{\cellcolor[HTML]{FFFBFB}19.01}  \\ \cline{3-11} 
\cellcolor[HTML]{F4F6F8}                                & \cellcolor[HTML]{F4F6F8}                            & \multicolumn{1}{r|}{\cellcolor[HTML]{F4F6F8}32}                                    & \multicolumn{1}{r|}{\cellcolor[HTML]{21629D}0.99}                       & \multicolumn{1}{r|}{\cellcolor[HTML]{2B69A2}1.23}                        & \multicolumn{1}{r|}{\cellcolor[HTML]{6D98BF}2.86}                        & \multicolumn{1}{r|}{\cellcolor[HTML]{FFFEFE}10.17}                       & \multicolumn{1}{r|}{\cellcolor[HTML]{FFF3F3}38.10}                       & \multicolumn{1}{r|}{\cellcolor[HTML]{FFEBEB}61.16}                        & \multicolumn{1}{r|}{\cellcolor[HTML]{FFA1A1}252.99}                       & \multicolumn{1}{r|}{\cellcolor[HTML]{FFEEEE}52.50}  \\ \cline{3-11} 
\cellcolor[HTML]{F4F6F8}                                & \cellcolor[HTML]{F4F6F8}                            & \multicolumn{1}{r|}{\cellcolor[HTML]{F4F6F8}64}                                    & \multicolumn{1}{r|}{\cellcolor[HTML]{2867A0}1.16}                       & \multicolumn{1}{r|}{\cellcolor[HTML]{447BAD}1.84}                        & \multicolumn{1}{r|}{\cellcolor[HTML]{A0BCD5}4.11}                        & \multicolumn{1}{r|}{\cellcolor[HTML]{FFFBFB}17.51}                       & \multicolumn{1}{r|}{\cellcolor[HTML]{FFE8E8}67.28}                       & \multicolumn{1}{r|}{\cellcolor[HTML]{FFD8D8}109.78}                       & \multicolumn{1}{r|}{\cellcolor[HTML]{FF5555}450.34}                       & \multicolumn{1}{r|}{\cellcolor[HTML]{FFDEDE}93.15}  \\ \cline{3-11} 
\cellcolor[HTML]{F4F6F8}                                & \cellcolor[HTML]{F4F6F8}                            & \multicolumn{1}{r|}{\cellcolor[HTML]{F4F6F8}128}                                   & \multicolumn{1}{r|}{\cellcolor[HTML]{3974A8}1.59}                       & \multicolumn{1}{r|}{\cellcolor[HTML]{5486B4}2.24}                        & \multicolumn{1}{r|}{\cellcolor[HTML]{CFDDEA}5.28}                        & \multicolumn{1}{r|}{\cellcolor[HTML]{FFFAFA}21.84}                       & \multicolumn{1}{r|}{\cellcolor[HTML]{FFE1E1}85.46}                       & \multicolumn{1}{r|}{\cellcolor[HTML]{FFCDCD}137.54}                       & \multicolumn{1}{r|}{\cellcolor[HTML]{FF2929}566.88}                       & \multicolumn{1}{r|}{\cellcolor[HTML]{FFD5D5}117.26} \\ \cline{3-11} 
\cellcolor[HTML]{F4F6F8}                                & \cellcolor[HTML]{F4F6F8}                            & \multicolumn{1}{r|}{\cellcolor[HTML]{F4F6F8}256}                                   & \multicolumn{1}{r|}{\cellcolor[HTML]{5285B3}2.21}                       & \multicolumn{1}{r|}{\cellcolor[HTML]{729CC1}3.00}                        & \multicolumn{1}{r|}{\cellcolor[HTML]{FAFCFD}6.35}                        & \multicolumn{1}{r|}{\cellcolor[HTML]{FFF8F8}25.19}                       & \multicolumn{1}{r|}{\cellcolor[HTML]{FFDDDD}96.91}                       & \multicolumn{1}{r|}{\cellcolor[HTML]{FFC7C7}154.43}                       & \multicolumn{1}{r|}{\cellcolor[HTML]{FF1010}630.75}                       & \multicolumn{1}{r|}{\cellcolor[HTML]{FFD0D0}131.26} \\ \cline{3-11} 
\cellcolor[HTML]{F4F6F8}                                & \multirow{-6}{*}{\cellcolor[HTML]{F4F6F8}HashedNet} & \multicolumn{1}{r|}{\cellcolor[HTML]{F4F6F8}512}                                   & \multicolumn{1}{r|}{\cellcolor[HTML]{83A8C9}3.42}                       & \multicolumn{1}{r|}{\cellcolor[HTML]{A7C1D8}4.28}                        & \multicolumn{1}{r|}{\cellcolor[HTML]{FFFFFF}8.06}                        & \multicolumn{1}{r|}{\cellcolor[HTML]{FFF7F7}27.91}                       & \multicolumn{1}{r|}{\cellcolor[HTML]{FFDADA}104.03}                      & \multicolumn{1}{r|}{\cellcolor[HTML]{FFC3C3}164.94}                       & \multicolumn{1}{r|}{\cellcolor[HTML]{FF0303}665.29}                       & \multicolumn{1}{r|}{\cellcolor[HTML]{FFCCCC}139.70} \\ \cline{2-11} 
\cellcolor[HTML]{F4F6F8}                                & \cellcolor[HTML]{F4F6F8}                            & \multicolumn{1}{r|}{\cellcolor[HTML]{F4F6F8}4}                                     & \multicolumn{1}{r|}{\cellcolor[HTML]{1E609C}0.92}                       & \multicolumn{1}{r|}{\cellcolor[HTML]{1E609C}0.92}                        & \multicolumn{1}{r|}{\cellcolor[HTML]{21629D}0.98}                        & \multicolumn{1}{r|}{\cellcolor[HTML]{4279AC}1.79}                        & \multicolumn{1}{r|}{\cellcolor[HTML]{99B7D2}3.94}                        & \multicolumn{1}{r|}{\cellcolor[HTML]{E5ECF3}5.82}                         & \multicolumn{1}{r|}{\cellcolor[HTML]{FFF9F9}22.39}                        & \multicolumn{1}{r|}{\cellcolor[HTML]{CEDCE9}5.25}   \\ \cline{3-11} 
\cellcolor[HTML]{F4F6F8}                                & \cellcolor[HTML]{F4F6F8}                            & \multicolumn{1}{r|}{\cellcolor[HTML]{F4F6F8}32}                                    & \multicolumn{1}{r|}{\cellcolor[HTML]{1F619D}0.95}                       & \multicolumn{1}{r|}{\cellcolor[HTML]{22639E}1.00}                        & \multicolumn{1}{r|}{\cellcolor[HTML]{2868A1}1.17}                        & \multicolumn{1}{r|}{\cellcolor[HTML]{4078AB}1.75}                        & \multicolumn{1}{r|}{\cellcolor[HTML]{A6C0D8}4.28}                        & \multicolumn{1}{r|}{\cellcolor[HTML]{F6F9FB}6.25}                         & \multicolumn{1}{r|}{\cellcolor[HTML]{FFF9F9}22.77}                        & \multicolumn{1}{r|}{\cellcolor[HTML]{D6E2ED}5.45}   \\ \cline{3-11} 
\cellcolor[HTML]{F4F6F8}                                & \cellcolor[HTML]{F4F6F8}                            & \multicolumn{1}{r|}{\cellcolor[HTML]{F4F6F8}64}                                    & \multicolumn{1}{r|}{\cellcolor[HTML]{3B75A9}1.62}                       & \multicolumn{1}{r|}{\cellcolor[HTML]{2767A0}1.15}                        & \multicolumn{1}{r|}{\cellcolor[HTML]{2C6AA2}1.26}                        & \multicolumn{1}{r|}{\cellcolor[HTML]{477DAE}1.92}                        & \multicolumn{1}{r|}{\cellcolor[HTML]{ADC5DB}4.45}                        & \multicolumn{1}{r|}{\cellcolor[HTML]{FEFEFE}6.45}                         & \multicolumn{1}{r|}{\cellcolor[HTML]{FFF9F9}24.01}                        & \multicolumn{1}{r|}{\cellcolor[HTML]{E6EDF4}5.84}   \\ \cline{3-11} 
\cellcolor[HTML]{F4F6F8}                                & \cellcolor[HTML]{F4F6F8}                            & \multicolumn{1}{r|}{\cellcolor[HTML]{F4F6F8}128}                                   & \multicolumn{1}{r|}{\cellcolor[HTML]{316EA4}1.38}                       & \multicolumn{1}{r|}{\cellcolor[HTML]{336FA5}1.44}                        & \multicolumn{1}{r|}{\cellcolor[HTML]{3772A7}1.52}                        & \multicolumn{1}{r|}{\cellcolor[HTML]{5587B4}2.26}                        & \multicolumn{1}{r|}{\cellcolor[HTML]{BFD2E3}4.90}                        & \multicolumn{1}{r|}{\cellcolor[HTML]{FFFFFF}7.25}                         & \multicolumn{1}{r|}{\cellcolor[HTML]{FFF8F8}27.18}                        & \multicolumn{1}{r|}{\cellcolor[HTML]{FFFFFF}6.56}   \\ \cline{3-11} 
\cellcolor[HTML]{F4F6F8}                                & \cellcolor[HTML]{F4F6F8}                            & \multicolumn{1}{r|}{\cellcolor[HTML]{F4F6F8}256}                                   & \multicolumn{1}{r|}{\cellcolor[HTML]{4C80B0}2.04}                       & \multicolumn{1}{r|}{\cellcolor[HTML]{4E82B1}2.10}                        & \multicolumn{1}{r|}{\cellcolor[HTML]{5083B2}2.14}                        & \multicolumn{1}{r|}{\cellcolor[HTML]{6D98BE}2.85}                        & \multicolumn{1}{r|}{\cellcolor[HTML]{D9E4EE}5.53}                        & \multicolumn{1}{r|}{\cellcolor[HTML]{FFFFFF}7.91}                         & \multicolumn{1}{r|}{\cellcolor[HTML]{FFF8F8}26.98}                        & \multicolumn{1}{r|}{\cellcolor[HTML]{FFFFFF}7.08}   \\ \cline{3-11} 
\multirow{-13}{*}{\cellcolor[HTML]{F4F6F8}sgd}          & \multirow{-6}{*}{\cellcolor[HTML]{F4F6F8}ROAST}     & \multicolumn{1}{r|}{\cellcolor[HTML]{F4F6F8}512}                                   & \multicolumn{1}{r|}{\cellcolor[HTML]{89ACCB}3.56}                       & \multicolumn{1}{r|}{\cellcolor[HTML]{7BA2C5}3.20}                        & \multicolumn{1}{r|}{\cellcolor[HTML]{81A6C7}3.36}                        & \multicolumn{1}{r|}{\cellcolor[HTML]{A2BED6}4.18}                        & \multicolumn{1}{r|}{\cellcolor[HTML]{FFFFFF}7.10}                        & \multicolumn{1}{r|}{\cellcolor[HTML]{FFFEFE}9.17}                         & \multicolumn{1}{r|}{\cellcolor[HTML]{FFF6F6}31.20}                        & \multicolumn{1}{r|}{\cellcolor[HTML]{FFFFFF}8.82}   \\ \hline
\end{tabular}
\caption{Total training step time  for different shapes of square weight matrix with input batch of 512. The tile-parameters of multiplication are optimized for each function over "forward +  backward" pass .The measurements are taken with tf32 on A100 (48GB)}
\label{tab:total-total}
\end{table}

\section{Variance in quality over different runs} \label{sec:variance}
\begin{figure}[h]
    \centering
    \begin{subfigure}[t]{0.45\textwidth}
    \includegraphics[scale=0.4]{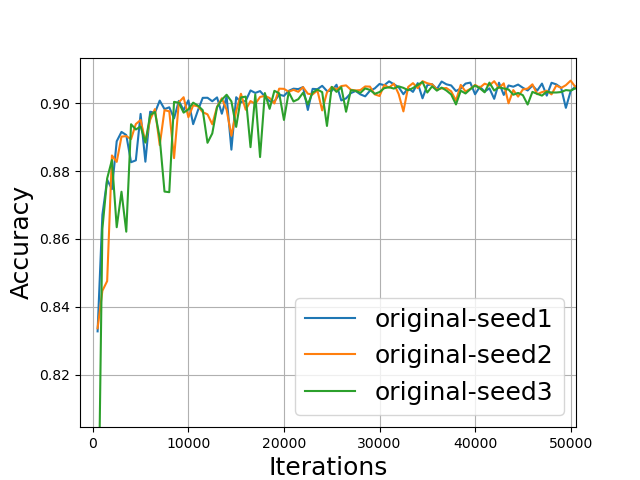}
    \caption{yelp - original}
    \centering
    \end{subfigure}
    \begin{subfigure}[t]{0.45\textwidth}
    \includegraphics[scale=0.4]{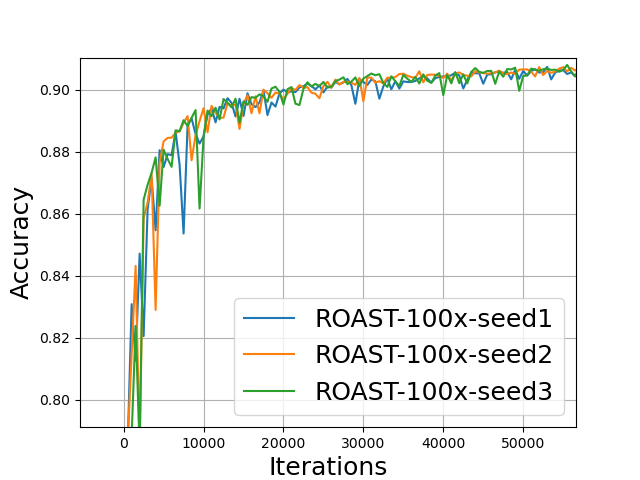}
    \caption{yelp - ROAST100x}
    \end{subfigure}
    \centering
    \begin{subfigure}[t]{0.45\textwidth}
\includegraphics[scale=0.4]{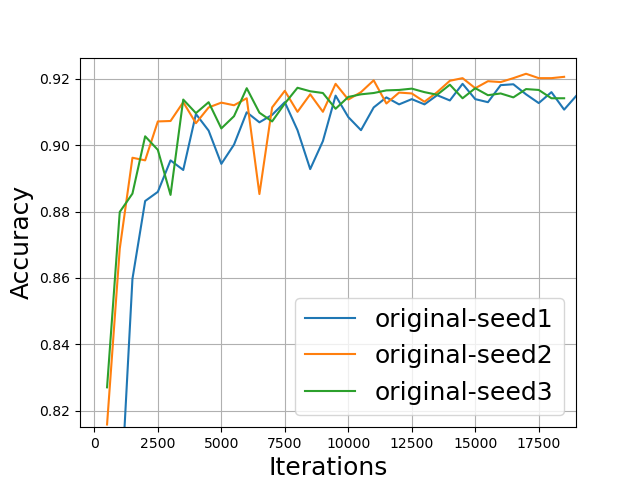}
    \caption{ag\_news - original}
    \end{subfigure}
    \centering
    \begin{subfigure}[t]{0.45\textwidth}
    \includegraphics[scale=0.4]{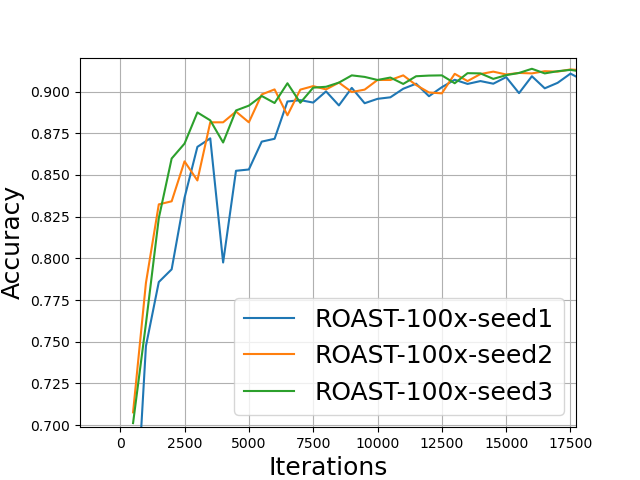}
    \caption{ag\_news - ROAST100x}
    \end{subfigure}
    \caption{Three runs of original and \roast-100x runs}
    \label{fig:variance}
\end{figure}

The figure~\ref{fig:variance} shows three runs of {\roast}ed BERT and BERT models

\end{document}